\documentclass{article}

\usepackage{microtype}
\usepackage{graphicx}
\usepackage{subcaption}
\usepackage{booktabs} 
\usepackage{pifont} 
\usepackage[table,xcdraw]{xcolor}
\usepackage{bm}
\usepackage{rotfloat}
\usepackage{rotating}
\usepackage{adjustbox}
\newcommand*\rot{\rotatebox{70 }}
\usepackage{marvosym}

\usepackage{hyperref}
\usepackage{ragged2e}
\usepackage{multirow}

\usepackage{amsmath}

\usepackage[preprint]{icml2026}

\usepackage{amsmath}
\usepackage{amssymb}
\usepackage{mathtools}
\usepackage{amsthm}
\definecolor{myblue}{RGB}{210, 225, 255}
\newcommand{\best}[1]{\textbf{\textcolor{red}{#1}}}
\newcommand{\secondbest}[1]{\underline{\textcolor{blue}{#1}}}

\usepackage[capitalize,noabbrev]{cleveref}

\theoremstyle{plain}

\theoremstyle{definition}

\theoremstyle{remark}

\usepackage[textsize=tiny]{todonotes}

\icmltitlerunning{GenShield: Unified Detection and Artifact Correction for AI-Generated Images}

\begin{document}

\twocolumn[
  \icmltitle{GenShield: Unified Detection and Artifact Correction for AI-Generated Images}



  \icmlsetsymbol{equal}{*}
  \icmlsetsymbol{leader}{${\ddagger}$}
  \icmlsetsymbol{intern}{${\dagger}$}
  \icmlsetsymbol{corresponding}{\scalebox{1.2}{\raisebox{0.1ex}{\Letter}}}

  \begin{icmlauthorlist}
    \icmlauthor{Zhipei Xu}{pku,tx,equal,intern}
    \icmlauthor{Xuanyu Zhang}{pku,equal}
    \icmlauthor{Youmin Xu}{tx,equal} \\
    \icmlauthor{Qing Huang}{pku} 
    \icmlauthor{Shen Chen}{tx,leader}
    \icmlauthor{Taiping Yao}{tx}
    \icmlauthor{Shouhong Ding}{tx}
    \icmlauthor{Jian Zhang}{pku,corresponding} \\
    
  \end{icmlauthorlist}

  \icmlaffiliation{pku}{School of Electronic and Computer Engineering, Peking University}
  \icmlaffiliation{tx}{Tencent Youtu Lab}

  \icmlcorrespondingauthor{Jian Zhang}{zhangjian.sz@pku.edu.cn}

    \icmlkeywords{Machine Learning, ICML}

  \vskip 0.3in
]




\printAffiliationsAndNotice{$^*$Equal contribution. $^{\dagger}$Work done during internship at Tencent Youtu Lab. $^{\ddagger}$Project Leader. $^{\scalebox{1.0}{\raisebox{0ex}{\Letter}}}$Corresponding author.}  

\begin{abstract}
Diffusion-based image synthesis has made AI-generated images (AIGI) increasingly photorealistic, raising urgent concerns about authenticity in applications such as misinformation detection, digital forensics, and content moderation. Despite the substantial advances in AIGI detection, how to correct detected AI-generated images with visible artifacts and restore realistic appearance remains largely underexplored. Moreover, few existing work has established the connection between AIGI detection and artifact correction. To fill this gap, we propose GenShield, a unified autoregressive framework that jointly performs explainable AIGI detection and controllable artifact correction in a closed loop from diagnosis to restoration, revealing a mutually reinforcing relationship between these two tasks. We further introduce a Visual Chain-of-Thought based curriculum learning  strategy that enables self-explained, multi-step ``diagnose-then-repair'' correction with an explicit stopping criterion. A high-quality dataset with large-scale ``artifact-restored'' pairs is also constructed alongside a unified evaluation pipeline. Extensive experiments on our correction benchmark and mainstream AIGI detection benchmarks demonstrate state-of-the-art performance and strong generalization of our method. The code is available at \url{https://github.com/zhipeixu/GenShield}.
\end{abstract}

\section{Introduction}


Recent advances in generative models, particularly diffusion-based image synthesis~\cite{wu2025qwenimagetechnicalreport,labs2025flux1kontextflowmatching,liu2025step1x-edit,seedream2025seedream,comanici2025gemini}, have enabled the creation of highly realistic images that are increasingly difficult to distinguish from real photographs. As AI-generated images (AIGI) are rapidly permeating social media, journalism, digital art, and online commerce, concerns regarding authenticity, trust, and visual credibility have become more prominent than ever.
The widespread adoption of image generation models has fundamentally altered the visual content ecosystem, where AIGI and camera-captured photographs now coexist at unprecedented scale. In such a setting, the ability to reliably determine whether an image is generated by an image generative model or captured by a real camera is no longer merely a technical challenge, but a critical requirement for a wide range of real-world applications, including misinformation detection, digital forensics, and content moderation.


AI-generated image detection~\cite{yan2024sanity,zhou2025aigi,wen2025spot} aims to determine whether an image has been produced by an image generator, typically by leveraging statistical cues, pixel-level artifacts, and semantic-level anomalies. 
A related and more fine-grained problem is synthetic image artifact correction~\cite{wang2025generated,fang2024humanrefiner}, which focuses on identifying and localizing non-natural regions and then correcting them to restore a more realistic appearance, such as structural inconsistencies, violations of physical laws, and local distortions. 
However, most existing studies~\cite{kang2025legion,fang2024humanrefiner,wang2025generated,shao2025finephys} primarily emphasize more accurate detection and localization, and commonly follow a pipeline that highlights artifacts with a bounding box or mask and applies a frozen inpainting model for local repainting. 
This design has several limitations: (i) the correction quality heavily depends on precise localization, which can be unreliable in practice; (ii) a frozen inpainting model becomes a performance bottleneck and limits the upper bound of correction; and (iii) inpainting often produces seams or inconsistencies between the repainted region and surrounding context, potentially introducing new artifacts. As a result, mask-free, end-to-end artifact correction remains largely underexplored. Moreover, existing datasets~\cite{kang2025legion} are largely detection-oriented: they provide artifact localization and textual descriptions, but rarely include paired restoration targets, which limits the progress of artifact correction methods.

Most existing studies overlook the potential synergy between correction and detection, even though \textbf{these two processes can naturally reinforce each other when modeled jointly}.
On the one hand, incorporating detection enables accurate anomaly identification and fine-grained artifact localization, which provides essential guidance for correction. This guidance steers the model to focus on truly problematic regions and remove non-natural cues without unnecessary changes. On the other hand, incorporating correction introduces a strong generative prior, strengthening the model’s ability to reconstruct realistic images and thereby sharpening its sensitivity to subtle artifacts that are otherwise hard to distinguish.
Therefore, it is natural to frame the task as a joint anomaly detection and repair problem, where the model not only detects and localizes artifacts but also corrects them to restore visual authenticity. Crucially, this setting reveals a mutually reinforcing relationship between understanding and generation, as diagnostic understanding guides targeted correction, and generation-based correction improves sensitivity for detection.

In recent years, the field of unified multimodal understanding and generation has witnessed rapid development. For example, Emu3~\cite{wang2024emu3} tokenizes images, text, and videos into a shared discrete space and trains a single transformer via next-token prediction. Show-o~\cite{xie2024show} unifies autoregressive and discrete diffusion modeling within one transformer for mixed-modality tasks such as VQA and text-to-image generation. BAGEL~\cite{deng2025emerging} presents a mixture-of-transformer-experts (MoT) architecture that harmonizes autoregressive language models with rectified flow, utilizing a dual-encoder design to effectively balance high-level semantic understanding with high-fidelity image generation.
These unified architectures suggest that understanding and generation can boost each other.
In this context, detection and artifact correction exhibit a similar understanding--generation duality, providing strong support for jointly modeling them within a unified framework.


Based on the above analysis, we propose GenShield, a unified autoregressive framework built upon a MoT architecture.
GenShield consists of two specialized experts for AI-generated image detection and artifact correction, operating on a shared multimodal backbone with shared self-attention. 
Within this unified architecture, GenShield supports two tightly coupled tasks:
(i) structured AIGI detection with explanatory rationales, and
(ii) artifact correction, which can be performed either via instruction-guided editing or through an iterative visual chain-of-thought (VCoT) diagnose--then--correct process that alternates between diagnostic reasoning and targeted image refinement.
To effectively train the model, we design a two-stage curriculum learning strategy.
Stage~1 focuses on instruction-guided correction to establish stable generative priors, while Stage~2 introduces multi-step VCoT self-correction. Throughout both stages, the AIGI detection task remains active, enabling joint optimization of detection and correction. Our contributions are summarized as follows.


\vspace{1pt} 
\noindent \ding{113}~(1) We propose GenShield, a unified autoregressive framework that connects AI image detection and artifact correction. By coupling semantic understanding with pixel-level reconstruction, our method forms an end-to-end loop from artifact diagnosis to authenticity restoration, showing the synergistic effects between detection and correction. 


\noindent \ding{113}~(2) We design a VCoT-based curriculum learning strategy.
By transitioning from instruction-guided correction to multi-step self-correction with an explicit stopping criterion, while keeping AIGI detection active throughout training, we establish a ``diagnose--then--correct'' paradigm that reduces learning complexity and improves logical transparency.

\vspace{1pt} 
\noindent \ding{113}~(3) We construct a specialized, high-quality dataset GenShield-Set tailored for unitied AIGI detection and correction. By leveraging explicit defect descriptions to guide advanced editors, we generate large-scale and precisely aligned ``artifact-restored'' image pairs. This fills a critical data gap and enables models to learn the mapping from synthetic anomalies to realistic images.

\vspace{1pt} 
\noindent \ding{113}~(4) Experimental results demonstrate that our GenShield achieves state-of-the-art performance on AIGI detection tasks, while its artifact correction capability surpasses that of advanced closed-source generators.





\begin{figure*}[t!]
	\centering
    \includegraphics[width=0.95\linewidth]{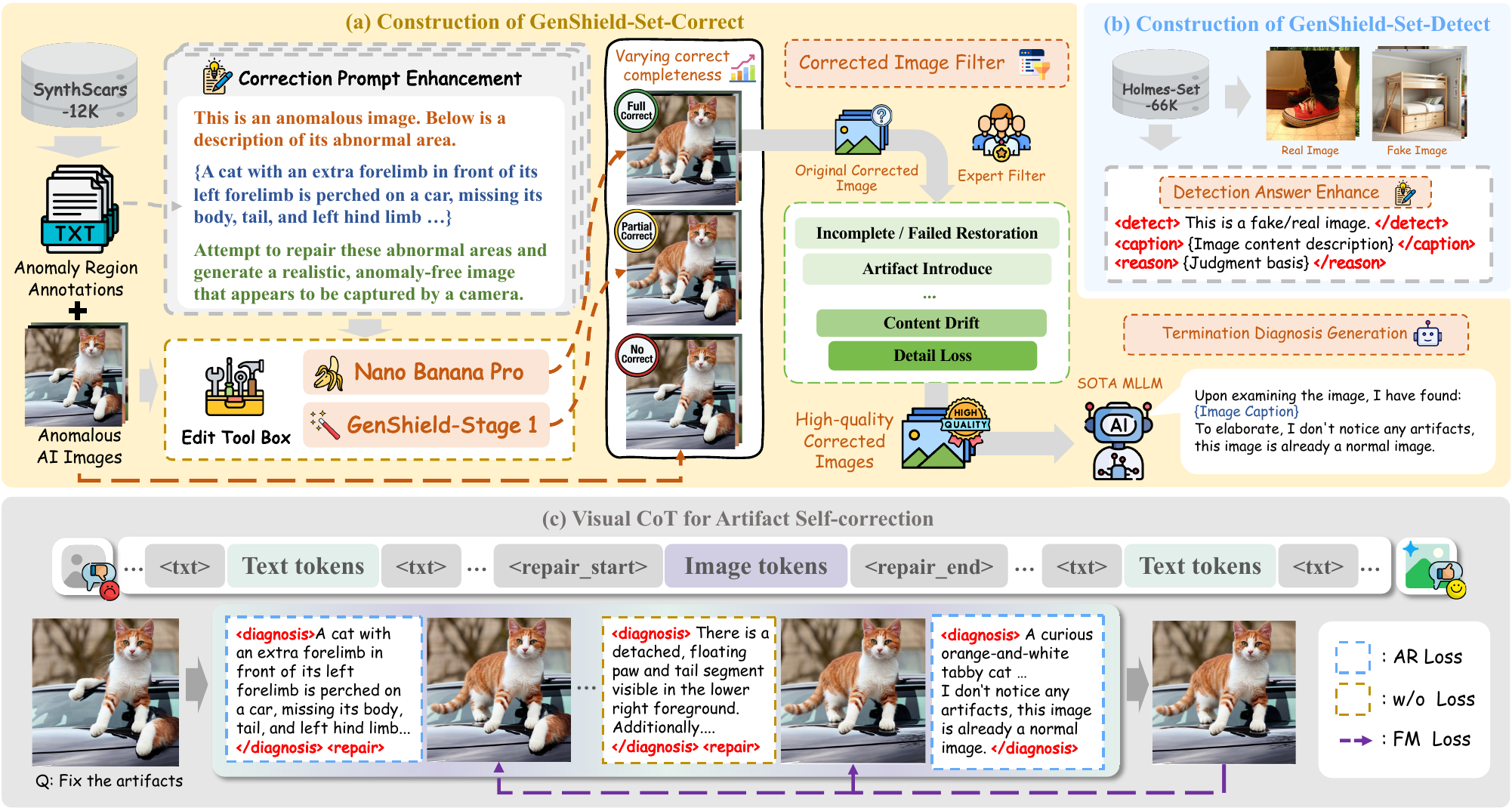}
	\vspace{-5pt}
	\caption{\textbf{Overview of GenShield-Set Construction.} (a) Correction data generation with prompt enhancement, expert filtering and termination answer generation. (b) Detection data enhancement with structured answers. (c) Interleaved text–image VCoT sequences for iterative diagnosis and correction.}
	\label{dataset}
    \vspace{-15pt}
\end{figure*}

\section{Related Works}

\subsection{Synthetic Image Detection}

Early synthetic image detection methods~\cite{wang2019cnnspot,zhong2023patchcraft,frank2020fredect,tan2023learning,wang2023dire} mainly use methods such as frequency domain artifacts, reconstruction residuals, and local pixel correlation to mine the low-level forgery clues introduced in the generation process.
With the rise of vision foundation models~\cite{radford2021learning,zhang2022dino}, semantic-level cues are increasingly adopted for detection. CLIP-based methods such as UniFD~\cite{ojha2023fakedetect}, C2P-CLIP~\cite{tan2025c2p} and FatFormer~\cite{liu2024forgery}  exploit image-text contrastive features to improve generalization, while self-supervised models like DINO offer transferable visual representations without relying on text supervision~\cite{guillaro2025bias,caitowards}.
Recently, image forgery detection has increasingly adopted LLMs to overcome the limitations of traditional black-box classifiers in generalization and explainability~\cite{xu2024fakeshield,liu2024forgerygpt,xu2025avatarshield,zhou2025aigi,kang2025legion}. By projecting image content into the language space, LLMs produce reasoning-rich text embeddings that enhance sensitivity to semantic anomalies. 
However, most existing methods still treat AIGI detection as an isolated task, optimizing accuracy, robustness, or explanations in separation. In contrast, the synergy between artifact correction and detection remains underexplored. Jointly modeling this complementarity is a promising yet insufficiently studied direction.


\subsection{Synthetic Image Artifact Localization and Correction}

Beyond global real/fake classification, recent studies have begun to investigate fine-grained anomaly understanding in synthetic images by explicitly characterizing and localizing non-natural regions. SynArtifact~\cite{cao2024synartifact} provides a taxonomy of 13 defect types and leverages vision–language models to both classify and localize these artifacts, enabling more structured diagnosis of generation-induced defects. AnomAgent~\cite{tan2025semantic} further targets semantic anomalies and proposes a multi-agent reasoning framework to identify and explain violations of physical laws or commonsense, emphasizing interpretable, human-like forensic analysis. In parallel, HumanRefiner~\cite{fang2024humanrefiner} narrows the scope to synthesized humans and focuses on biological plausibility, using skeleton pose priors to detect anatomical deformities, and then relies on a frozen generator to perform regeneration or inpainting based on the detected results.
While these efforts advance artifact categorization, localization, and explanation, they largely emphasize anomaly diagnosis rather than end-to-end correction, leaving the joint modeling of detection-driven diagnosis and controllable repair an open direction.

\subsection{Unified Understanding and Generation}
Compared with conventional pipelines that treat image understanding and generation as separate processes, recent work increasingly unifies them within a single framework, where joint optimization improves overall capability and generalization.
Inspired by language modeling, some studies~\cite{chen2025janus,wang2024emu3,wu2025janus} encode images into discrete tokens and generate them autoregressively via next-token prediction. Other approaches~\cite{xie2024show,zhou2024transfusion} incorporate diffusion mechanisms, using text-token generation as a bridge to model continuous images and improve generation quality. Another line of work~\cite{lin2025uniworld,chen2025blip3} emphasizes modular designs, decoupling perception and generation to better optimize each component. The recent BAGEL model~\cite{deng2025emerging} adopts a Mixture-of-Transformers (MoT) design, separating parameters for understanding and generation while sharing self-attention, further boosting performance. 
This surge of unified understanding-and-generation models makes the synergy between AIGI detection and artifact correction both feasible and compelling. 
\vspace{-10pt}

\begin{figure*}[t!]
	\centering
    \includegraphics[width=0.95\linewidth]{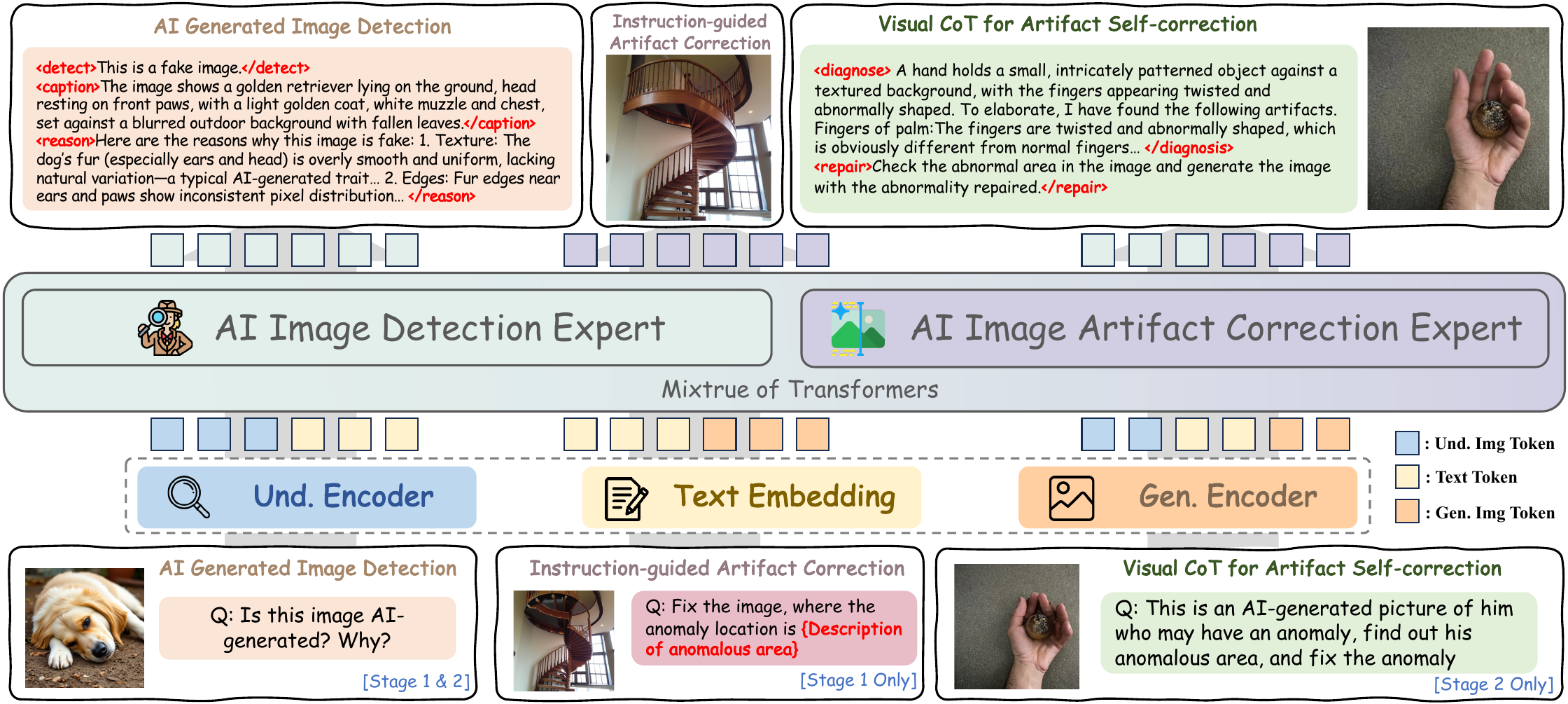}
	\vspace{-5pt}
    \caption{\textbf{Pipeline of GenShield.} Stage~1 performs AI-generated image detection and instruction-guided artifact correction, while Stage~2 extends correction to an iterative VCoT setting, alternating between multi-step diagnosis and refinement.}
	\label{pipeline}
    \vspace{-16pt}
\end{figure*}

\section{Methodology}

\subsection{Construction of GenShield-Set}

As shown in Fig.~\ref{dataset}, we construct GenShield-Set for joint training of explainable AIGI detection and artifact correction, comprising GenShield-Set-Correct with paired correction targets and GenShield-Set-Detect with structured detection answers.

\textbf{GenShield-Set-Correct.} As illustrated in Fig.~\ref{dataset}(a), we build GenShield-Set-Correct on top of SynthScars~\cite{kang2025legion}, which provides anomalous synthetic images together with textual anomaly-region annotations. For each anomalous image $\textbf{I}_{\text{artifact}}$ and its annotation $\textbf{T}_\text{diag}$, we first perform correction prompt enhancement, rewriting the annotation into a standardized instruction . Using Nano Banana Pro~\cite{comanici2025gemini}, the most advanced image editor, we generate the corrected candidates. 
Since automatic editing may result in under-repair or over-editing, we apply a human expert-based corrected image filter to select high-quality restorations $\textbf{I}_{\text{correct}}$. This filter removes candidates with common failure modes, including incomplete or failed restoration, newly introduced artifacts, content drift, and excessive detail loss, ensuring that the retained repaired images are both artifact-reduced and semantically consistent with the originals.
Subsequently, we additionally create a termination diagnosis $\textbf{T}_{\text{stop}}$ for $\textbf{I}_{\text{correct}}$ by prompting Qwen-2.5-VL~\cite{Qwen2.5-VL} to explicitly state that no artifacts are observed and the image already appears normal. This termination diagnosis is later used in our iterative VCoT process to determine when further correction is unnecessary. 
Notably, to guide multi-round iterative correction, we use the intermediate results of GenShield (in training stage 1) to generate a set of medium-quality corrected images $\textbf{I}_{\text{mid}}^{(i)}$. These predictions typically apply relatively mild edits and therefore only partially remove the artifacts in the original images.
Finally, we obtain over 10K high-quality, precisely aligned tuples 
$(\textbf{I}_{\text{artifact}}, \textbf{T}_{\text{diag}}, \textbf{I}^{(1)}_{\text{mid}}, \ldots,\textbf{I}_{\text{mid}}^{(N)},\textbf{I}_{\text{correct}}, \textbf{T}_{\text{stop}})$ as training data. More details are provided in the \textbf{\textit{Appendix~\ref{app:Details of GenShield-Set}}}.


\textbf{GenShield-Set-Detect.}
As illustrated in Fig.~\ref{dataset}(b), to train the detection expert, we construct GenShield-Set-Detect primarily from Holmes-Set~\cite{zhou2025aigi}, which contains real and AI-generated image with detailed explanatory detection annotations. We further perform detection answer enhancement by converting the raw annotations into a unified, structured format that includes an authenticity prediction (\texttt{<detect>}), an image caption (\texttt{<caption>}), and an evidence-based justification (\texttt{<reason>}). Finally, we obtain 66K image--text pairs for training.

\subsection{Overview of GenShield}

\textbf{Motivation.} Nowadays, with the growing synergy between multimodal understanding and image generation~\cite{deng2025emerging,xie2024show}, we explore whether AI image detection can be jointly improved with a complementary generative task. 
We observe that artifact correction is highly aligned with AIGI detection: detection produces structured and fine-grained diagnostic cues that can guide targeted correction, while correction encourages the model to reconstruct realistic distributions, thereby improving its sensitivity and generalization to subtle artifacts. 
However, existing pipelines~\cite{fang2024humanrefiner,wang2025generated,zhou2025aigi,kang2025legion} typically decouple these two processes or only focus on a single task, which limits controllability and hinders mutual reinforcement.

\textbf{Task Definition.} Motivated by the complementary nature of explainable detection and artifact correction, we introduce GenShield, a unified autoregressive framework that connects AIGI detection with controllable restoration in a single model, as illustrated in Fig.~\ref{pipeline}. 
Given an input image and a task instruction, GenShield supports two tightly coupled tasks: 
(i) \textit{AIGI detection}, where the model autoregressively generates a structured decision together with forensic rationales; and 
(ii) \textit{artifact correction}, which can be carried out either via instruction-guided editing under explicit artifact descriptions or through an iterative VCoT diagnose--then--correct process that alternates between diagnostic reasoning and targeted image refinement with an explicit stopping criterion.

\textbf{Model Architecture.} GenShield adopts a mixture-of-transformers~\cite{deng2025emerging} design with two specialized experts, an AI Image Detection Expert and an AI Image Artifact Correction Expert, built on top of a shared decoder-style multimodal backbone. Both experts operate on the same multimodal token sequence and interact through shared multi-modal self-attention at every layer, which serves as the key mechanism for mutual reinforcement. On one hand, the correction expert injects camera-consistent generative priors into the shared backbone, aligning detection features with real-image statistics and sharpening sensitivity to subtle artifacts. On the other hand, during correction, the detection expert provides repair-oriented diagnostic descriptions that are propagated through the same attention pathway as explicit guidance, enabling more precise edits with better semantic preservation. Together, shared self-attention enables bidirectional information flow between understanding and generation, forming the architectural basis of GenShield’s closed-loop synergy between detection and correction.

\subsection{VCoT-based Curriculum Learning}

Synthetic artifact correction must satisfy two competing requirements: accurately identifying and removing non-natural artifacts, while preserving identity, semantics, and global structure. 
Directly learning an end-to-end mapping from ``anomalous image → refined image'' often suffers from sparse supervision, uncontrolled edit scope, and semantic drift. 
To address this, we propose a VCoT-based curriculum learning strategy, which explicitly decomposes correction into a two-step reasoning pipeline, Diagnose-then-Repair, and progressively builds reliable diagnosis and controllable repair via an easy-to-hard training schedule.
In addition, we keep the AIGI detection task active throughout training, allowing detection and correction to mutually reinforce each other.



\paragraph{Detection Part.}
We formulate AIGI detection as autoregressive structured-text generation. Given the image prompt context
$\mathbf{x}_{\text{cond}}=[\textbf{I}_{\text{det}}, \textbf{T}_{\text{det}}]$ with length $l_{\text{con}}$, the model generates the detection answer
as text tokens appended to the context. We train the detection expert by maximum likelihood with an autoregressive objective:


\vspace{-20pt}
\begin{equation}
\scalebox{0.99}{$
\mathcal{L}_{AR}(\theta) = -\mathbb{E}_{(\mathbf{x}_{\text{cond}},\mathbf{x}_{\text{text}})}\left[\sum_{i=l_\text{con}}^{l-1} \text{log} ~ \text{P}_{\theta}(\mathbf{x}_{i+1}|\mathbf{x}_{1:i}) \right]$.}
\end{equation}
\vspace{-20pt}

Here $\mathbf{x}_{\text{text}}$ denotes the output text,  here is the detection results and explaination. $l$ is the total sequence length. The prefix $\mathbf{x}_{1:i}$ contains all previous multimodal tokens. This objective supervises the model to generate structured detection outputs that include both an authenticity decision and evidence-based rationales.

\paragraph{Correction Part.}
We decompose the optimization of the correction component into an easy-to-hard curriculum with two stages: instruction-guided correction and Visual CoT for self-correction.

\begin{table*}[t]
    \centering
    \renewcommand{\arraystretch}{1.2}
    \caption{Evaluation of AI synthetic image detection ability on Holmes-Set~\cite{zhou2025aigi}. \best{Bold} indicates the best result, and \secondbest{underlined} denotes the second best.}
    \label{tab:detection}
    \vspace{-5pt}
    \resizebox{\textwidth}{!}{
        \begin{tabular}{lcccccccccccccccccccc|cc}
            \toprule
            \multirow{2}*{\textbf{Method}} & \multicolumn{2}{c}{\textbf{Janus}} & \multicolumn{2}{c}{\textbf{Janus-Pro-1B}} & \multicolumn{2}{c}{\textbf{Janus-Pro-7B}} & \multicolumn{2}{c}{\textbf{Show-o}} & \multicolumn{2}{c}{\textbf{LlamaGen}} & \multicolumn{2}{c}{\textbf{Infinity}} & \multicolumn{2}{c}{\textbf{VAR}} & \multicolumn{2}{c}{\textbf{PixArt-XL}} & \multicolumn{2}{c}{\textbf{SD3.5-Large}} & \multicolumn{2}{c|}{\textbf{FLUX}} & \multicolumn{2}{c}{\textbf{Mean}} \\ 

            \cmidrule(lr){2-3} \cmidrule(lr){4-5} \cmidrule(lr){6-7} \cmidrule(lr){8-9} \cmidrule(lr){10-11} \cmidrule(lr){12-13} \cmidrule(lr){14-15} \cmidrule(lr){16-17} \cmidrule(lr){18-19} \cmidrule(lr){20-21} \cmidrule(lr){22-23} 
            
            ~ & Acc. & A.P. & Acc. & A.P. & Acc. & A.P. & Acc. & A.P. & Acc. & A.P. & Acc. & A.P. & Acc. & A.P. & Acc. & A.P. & Acc. & A.P. & Acc. & A.P. & Acc. & A.P.  \\ 
            
            \midrule
            
            \multicolumn{5}{l}{\textit{\textbf{Non-LLM Based AI Image Detector}}} \\ 
            \midrule
            \quad CNNSpot & 70.0 & 86.0 & 70.9 & 85.8 & 85.0 & 93.6 & 72.2 & 86.0 & 61.9 & 71.4 & 86.8 & 94.6 & 59.9 & 75.0 & 78.2 & 90.1 & 63.8 & 81.1 & 79.9 & 92.0 & 72.9 & 85.6 \\
            \quad AntifakePrompt  & 72.2 & 87.4 & 84.3 & 94.0 & 84.8 & 93.1 & 86.2 & 95.5 & 96.2 & 99.4 & 83.6 & 94.1 & 90.7 & 95.6 & 81.7 & 92.8 & 92.8 & 97.8 & 66.1 & 80.8 & 83.9 & 93.1 \\
            \quad UniFD  & 87.6 & 97.8 & 96.9 & 99.5 & 96.4 & 99.5 & 85.9 & 97.4 & 93.1 & 98.6 & 79.2 & 96.2 & 64.3 & 85.9 & 75.7 & 94.4 & 87.8 & 97.8 & 69.6 & 91.4 & 83.6 & 95.9 \\
            \quad NPR  & 51.2 & 55.9 & 69.5 & 75.1 & 73.9 & 77.9 & 93.7 & 99.6 & 93.5 & 99.4 & 93.8 & 99.9 & 85.9 & 91.2 & 93.4 & 99.1 & 91.6 & 97.7 & 93.6 & 99.5 & 84.0 & 89.5 \\
            \quad LaRE  & 70.8 & 99.3 & 74.7 & 97.5 & 95.6 & 99.7 & 80.0 & 99.0 & 91.6 & 99.6 & 77.9 & 99.6 & 98.8 & 100.0 & 82.2 & 99.7 & 94.1 & 99.5 & 84.3 & 99.0 & 85.0 & 99.3 \\
            \quad RINE & 89.9 & 98.3 & 98.7 & 99.9 & 97.2 & 99.6 & \secondbest{98.8} & 99.9 & \secondbest{99.1} & \secondbest{100.0} & \secondbest{99.2} & 99.9 & 85.0 & 97.9 & 98.9 & 99.8 & 97.8 & 99.7 & \secondbest{97.1} & 99.7 & 96.2 & 99.5 \\
            \quad AIDE   & 91.2 & 99.1 & 98.9 & \secondbest{99.9} & \secondbest{97.8} & \secondbest{99.8} & 98.0 & 99.8 & \best{99.4} & \best{100.0} & 98.7 & \secondbest{99.9} & 93.6 & 99.3 & 98.6 & 99.9 & \best{99.4} & \best{100.0} & 94.4 & \secondbest{99.5} & \secondbest{97.0} & \secondbest{99.7} \\

            \midrule
            
            \multicolumn{11}{l}{\textit{\textbf{LLM Based AI Image Detector}}} \\ 
            \midrule

            \quad AIGI-Holmes & 95.7 & \secondbest{99.8} & 99.1 & \best{100.0} & 93.4 & 99.6 & 97.5 & \secondbest{99.9} & 98.0 & 99.9 & \best{99.6} & \best{100.0} & 99.2 & \best{100.0} & 95.2 & 99.8 & \secondbest{98.8} & \secondbest{99.9} & 79.5 & 96.5 & 95.6 & 99.5  \\
            \quad FakeVLM  & 86.8 & 93.0 & 83.3 & 83.5 & 68.0 & 72.6 & 78.3 & 81.9 & 86.8 & 87.3 & 83.4 & 85.3 & 82.0 & 84.1 & 77.1 & 82.3 & 82.5 & 84.2 & 83.7 & 89.5 & 81.2 & 84.4  \\
            \quad Qwen2.5-VL-7B  & \best{99.6} & \best{99.9} & \best{99.8} & 99.8 & 63.0 & 77.5 & 84.9 & 90.7 & 80.7 & 89.9 & 98.1 & 98.7 & \best{99.8} & 99.9 & \best{99.6} & \secondbest{99.8} & 98.2 & 98.9 & 54.2 & 66.2 & 87.8 & 92.1  \\
            \rowcolor{myblue}
            \quad \textbf{Ours} & \secondbest{98.8} & 99.7 & \secondbest{99.5} & 99.8 & \best{99.4} & \best{99.9} & \best{99.2} & \best{99.9} & 98.8 & 99.5 & 98.9 & 99.7 & \secondbest{99.4} & \secondbest{99.9} & \secondbest{99.3} & \best{99.9} & 96.7 & 99.8 & \best{97.9} & \best{99.7} & \best{98.8}  & \best{99.8}  \\  

            \bottomrule
        \end{tabular}
    }
    \vspace{-10pt}
\end{table*}

\emph{Stage 1: Instruction-guided correction.} 
In Stage~1, we train the correction expert with strong supervision using instruction-guided correction, which provides a stable learning signal before introducing iterative self-correction. 
Concretely, the input consists of the anomalous image $\textbf{I}_{\text{artifact}}$ and a correction instruction derived from the diagnostic annotation $\textbf{T}_{\text{diag}}$ (a detailed and structured artifact description), and the model learns the mapping $\mathbf{x}_{\text{cond}}=[ \textbf{I}_{\text{artifact}}, \textbf{T}_{\text{diag}}] \rightarrow \textbf{I}_{\text{correct}}$.
The model is optimized to generate the target corrected image $\mathbf{I}_{\text{correct}}$ in a flow matching manner.
\begin{equation}
\label{fm}
\scalebox{0.85}{$
\mathcal{L}_{FM}(\theta) = \mathbb{E}_{\mathbf{z}_0\sim\mathcal{N}(0,\mathbf{I})} 
 \left[\|v_{\theta}(\mathbf{z}_t, t| \mathbf{x}_{\text{cond}})-(\mathbf{I}_{\text{correct}}-\mathbf{z}_0)\|^2 \right]$,}
\end{equation}where $\mathbf{z}_t=t\mathbf{I}_\text{correct}+(1-t) \mathbf{z}_0$, $v_\theta$ denotes the velocity neural network used for artifact correction with parameter $\theta$, and $t$ is the sampled diffusion timestep. This stage encourages the model to learn a realistic correction prior and to perform localized, semantically preserving edits under explicit guidance, thereby reducing uncontrolled edit scope and semantic drift. 
The trained model also serves as the initialization for Stage~2, and is used to produce intermediate corrections $\textbf{I}_{\text{mid}}^{(n)}$ that diversify the starting states for multi-round VCoT training.

\emph{Stage 2: Visual CoT for self-correction.} 
As shown in Fig.~\ref{dataset}~(c), Stage~2 upgrades correction from externally guided editing to self-correction with multi-step Visual CoT. Starting from a simple prompt $\textbf{Q}$ (e.g. ``Please repair this image.'') and an input image $\mathbf{I}_{\text{artifact}}$, the model first generates a repair diagnosis $\hat{\textbf{T}}_{\text{diag}}$ that describes remaining suspicious regions and artifact cues, and then performs a correction step conditioned on this diagnosis to produce an updated image. 
The updated image is re-fed into the model to trigger the next diagnose--correct iteration, forming an alternating multi-round process that progressively removes residual artifacts. The process can be formulated as. 
\begin{equation}
\scalebox{0.8}{$
\underbrace{[\mathbf{Q}, \mathbf{I}_{\text{artifact}}]}_{\text{Initial state}}
\;\rightarrow\;
\underbrace{
[\mathbf{T}^{(1)}_{\text{mid}}, \mathbf{I}^{(1)}_{\text{mid}}]
\;\rightarrow\; \cdots \;\rightarrow\;
[\mathbf{T}^{(N)}_{\text{mid}}, \mathbf{I}^{(N)}_{\text{mid}}]
}_{\text{Intermediate state}}
\;\rightarrow\;
\underbrace{[\mathbf{T}_{\text{stop}}, \mathbf{I}_{\text{correct}}]}_{\text{Termination state}}
$} \notag
\end{equation}

\textcolor{blue}{\emph{Initial state.}} We adopt an interleaved image—text training, where a simple prompt is progressively refined into a detailed restoration instruction and simultaneously performing an initial correction of the image, namely $\mathbf{x}_{\text{cond}} = [\mathbf{Q}, \mathbf{I}_{\text{artifact}}] \rightarrow [\mathbf{T}_{\text{diag}}, \mathbf{I}_{\text{correct}}]$. The loss function is $\mathcal{L}=\mathcal{L}_{FM} + \lambda\mathcal{L}_{AR}$, $\lambda=0.25$. 

\textcolor{blue}{\emph{Intermediate state.}} To make this self-correction chain trainable, we supervise each produced image of the intermediate state toward the same target $\textbf{I}_{\text{correct}}$ via the loss function $\mathcal{L} = \mathcal{L}_{FM}$, namely $\mathbf{x}_{\text{cond}} = [\mathbf{Q}, \mathbf{I}^{(i)}_{\text{mid}}] \rightarrow \mathbf{I}_{\text{correct}}$. To be noted, we leave intermediate diagnostic texts $\mathbf{T}^{(i+1)}_{\text{mid}}$ unconstrained in later rounds to encourage free-form yet actionable reasoning. 

\textcolor{blue}{\emph{Terminate state.}} To explicitly instruct the model to determine when to stop restoration, the ground-truth $\mathbf{I}_{\text{correct}}$ is provided as the image input, such that the model preserves the image while generating the termination text $\mathbf{T}_{\text{stop}}$.
The loss function $\mathcal{L}=\mathcal{L}_{FM} + \lambda\mathcal{L}_{AR}$, with $\mathbf{x}_{\text{cond}} = [\mathbf{Q}, \mathbf{I}_{\text{correct}}] \rightarrow [\mathbf{T}_{\text{stop}}, \mathbf{I}_{\text{correct}}]$. During inference, when the model is fed with satisfactory results, our GenShield will output the text like `` I do not notice any artifacts,
this image is already a normal image'' and terminate the multi-step restoration process.
This design enables the model to automatically decide when further correction is unnecessary, leading to stable multi-step refinement without over-editing.
\vspace{-10pt}

\begin{table*}[t]
    \centering
    \renewcommand{\arraystretch}{1.2}
    \caption{Evaluation of AI-generated artifact correction ability on SynthScars~\cite{kang2025legion}. \best{Bold} and \secondbest{underline} denote the best and second best result. [$^*$: our single-step correction variant, $^\triangle$: LEGION equipped with an external SDXL inpainting module.]}

    \label{tab:refine}
    \vspace{-5pt}
    \resizebox{1.0\linewidth}{!}{
        \begin{tabular}{lccccccccccc} 
            \toprule
            \multirow{2}{*}{\textbf{Method}}  & \multicolumn{4}{c}{\textbf{GPT-assisted Evaluation}} & \multicolumn{4}{c}{\textbf{Human Evaluation}} & \multicolumn{3}{c}{\textbf{Objective Metrics}}\\ 
            
            \cmidrule(lr){2-5} \cmidrule(lr){6-9} \cmidrule(lr){10-12} 
            
            & Structure $\downarrow$ & Physics $\downarrow$ & Distortion $\downarrow$ & Mean $\downarrow$ & Structure $\downarrow$ & Physics $\downarrow$ & Distortion $\downarrow$ & Mean $\downarrow$ & HPSv3 $\uparrow$ & CLIP-Score $\uparrow$ & PickScore $\uparrow$  \\ 
            \midrule
            
            
            \multicolumn{11}{l}{\textit{\textbf{Closed-Sourced Image Edit Model}}} \\ 
            \midrule
            \quad GPT-Image & 0.24 & \secondbest{0.13} & 0.34 & 0.24 & 0.28 & 0.20 & 0.20 & 0.23  & 6.09 & 21.83 & 18.70
 \\ 
            \quad Nano-Banana-Pro & 0.22 & 0.21 & \secondbest{0.29} & 0.24 & 0.21 & 0.28 & 0.26 & 0.25 & 5.92 & 21.89 & 18.71  \\
            \quad Nano-Banana & 0.25 & 0.24 & 0.32 & 0.27 & 0.20 & \secondbest{0.18} & 0.27 & 0.21 & 5.77 & 21.40 & 18.64 \\ 
            \quad Seedream & 0.33 & 0.31 & 0.39 & 0.34 & 0.31 & 0.28 & 0.28 & 0.29 & 6.12 & 21.76 & 18.62 \\ 
            \quad FLUX-Pro & 0.27 & 0.33 & 0.29 & 0.30 & 0.21 & 0.29 & 0.20 & 0.23 & 5.98 & 22.07 & 18.68 \\
            
            \midrule
            
            \multicolumn{11}{l}{\textit{\textbf{Open-Sourced Image Edit Model}}} \\ 
            \midrule

            \quad BAGEL & 0.70 & 0.69 & 0.84 & 0.74 & 0.72 & 0.71 & 0.79 & 0.74 & 4.68 & 21.44 & 18.54 \\ 
            \quad Qwen-Image-Edit & 0.49 & 0.48 & 0.61 & 0.53 & 0.37 & 0.35 & 0.51 & 0.41 & 5.99 & 21.41 & 18.54 \\ 
            \quad Step1X-Edit & 0.43 & 0.37 & 0.48 & 0.43 & 0.39 & 0.36 & 0.41 & 0.39  & 5.47 & 21.33 & 18.60 \\ 
            \midrule
            
            \multicolumn{11}{l}{\textit{\textbf{Artifact Correction Model}}} \\ 
            \midrule

            \quad LEGION$^\triangle$ & 0.35 & 0.35 & 0.47 & 0.39  & 0.39 & 0.41 & 0.3 & 0.37  & 5.01 & 21.46 & 18.61 \\ 
            \rowcolor{myblue}
            \quad \textbf{Ours*} & \secondbest{0.18} & 0.21 & 0.30 & \secondbest{0.23}  & \secondbest{0.17} & 0.25 & \best{0.16} & \secondbest{0.19}  & \best{6.23} & \secondbest{22.11} &  \secondbest{18.83}  \\  
            \rowcolor{myblue}
            \quad \textbf{Ours} & \best{0.13} & \best{0.10} & \best{0.21} & \best{0.15} & \best{0.15} & \best{0.12} & \secondbest{0.19} & \best{0.16} & \secondbest{6.20} & \best{22.12} & \best{18.86}  \\  
            \bottomrule
        \end{tabular}
    }
    \vspace{-15pt}
\end{table*}

\section{Experiment}

\subsection{Experiment Setup}

%

\textbf{Implementation details.} We follow a unified multi-task training pipeline with a curriculum schedule. In Stage 1, we jointly train detection and instruction-guided repair. In Stage 2, we keep detection training unchanged while upgrading repair to the Visual CoT diagnose-then-repair formulation. 
We build our method on top of BAGEL~\cite{deng2025emerging}, which employs a ViT-style encoder~\cite{dosovitskiy2020image} for visual understanding and whose VAE is from FLUX~\cite{labs2025flux1kontextflowmatching}. 
We use AdamW in both stages with a fixed learning rate of $2\times10^{-5}$ and a 500-step warmup.
We freeze the Und. Encoder and Gen. Encoder, while keeping all other components trainable.
More details are provided in the \textbf{\textit{Appendix~\ref{app:More Implementation Details}}}.
\vspace{-5pt}


\subsection{AIGI Detection Performance Evaluation}

We evaluate our method on the Holmes-Set~\cite{zhou2025aigi}, which consists of AI-generated and real images across various generative models. To ensure a fair comparison, all methods were retrained on this dataset under identical conditions. We categorize the compared models into LLM-based and non-LLM-based AI image detectors based on whether their architecture incorporates a large language model (LLM). We use Acc. (accuracy) and A.P. (area under the precision-recall curve) as the detection evaluation metrics.

As shown in Table~\ref{tab:detection}, our method outperforms existing SOTA methods in both Acc. and A.P. across multiple generative models. Specially on Janus-Pro-7B, our approach achieves 99.4\% accuracy and 99.9\% A.P., far exceeding the second-best result of 97.8\% accuracy and 99.8\% A.P. achieved by AIDE. Our model also leads on the mean performance across all generators, with a 98.8\% accuracy and 99.8\% A.P..
The results demonstrate the strong performance of our method, which can be attributed to the synergistic integration of explainable detection and artifact correction within a unified framework. By leveraging both detailed diagnostic reasoning and robust generative priors, our approach effectively detects subtle anomalies, leading to its superior performance on the Holmes-Set benchmark.
Additionally, an interpretable subjective result, as shown in Fig.~\ref{zhuguan}(c), demonstrates how our model accurately detects that the image is fake and provides a detailed analysis of why it is considered fake from texture, shadows, and lighting.
More experimental results can be found in the  \textbf{\textit{Appendix~\ref{app:More Experimental Results}}}.

\subsection{Artifact Correction Performance Evaluation}
\label{sec:exp_refine}

To evaluate the artifact correction performance of our method, we evaluate it on the SynthScars~\cite{kang2025legion} benchmark. To ensure a comprehensive evaluation, we compare our model with advanced closed-sourced and open-sourced image edit methods and artifact correction methods. We assess performance using both subjective and objective evaluations, with results shown in Table~\ref{tab:refine}.

\begin{figure*}[t!]
	\centering
    \includegraphics[width=1.0\linewidth]{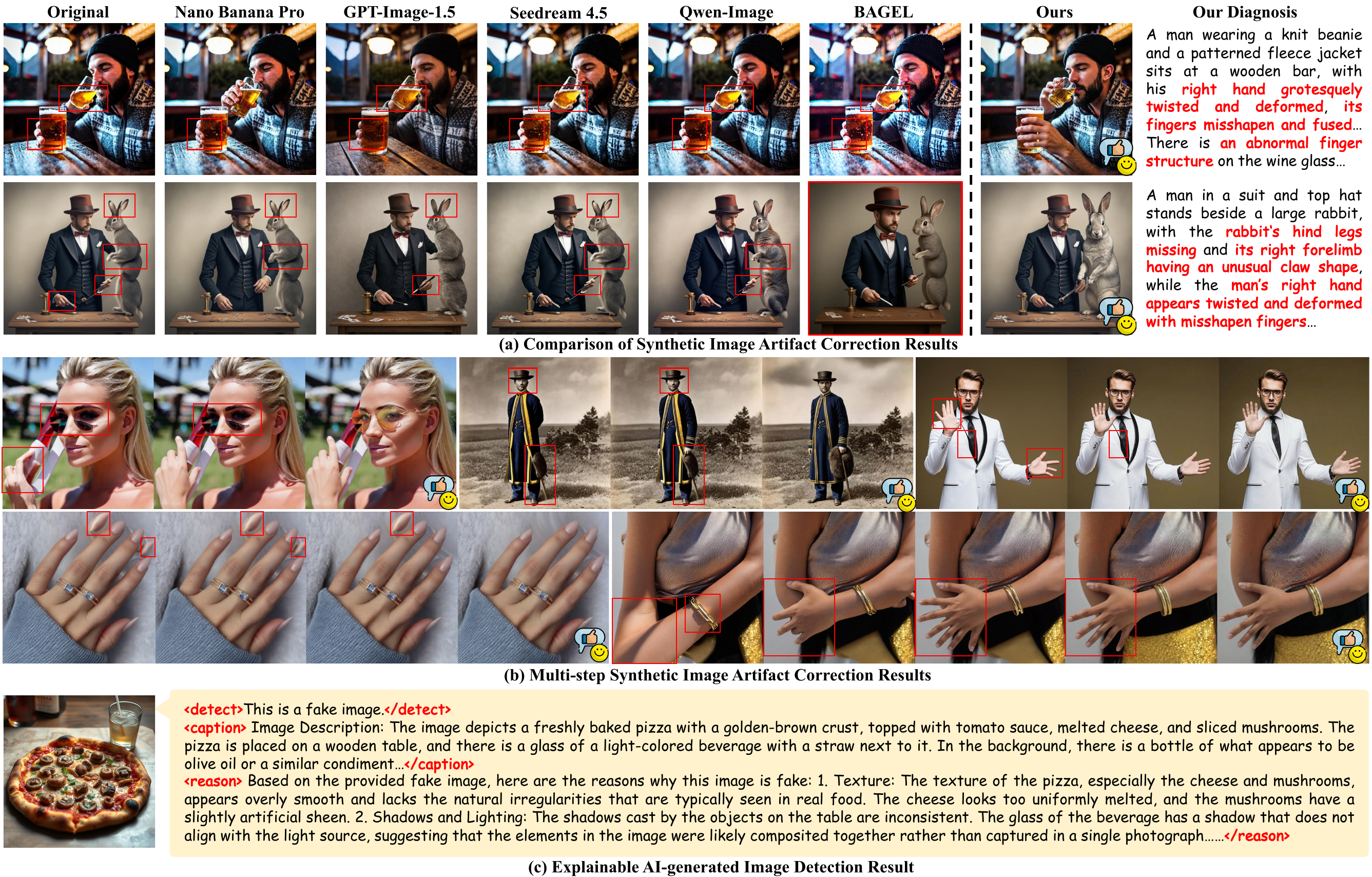}
	\vspace{-15pt}
	\caption{Artifact Correction and Synthetic Detection Results of our GenShield.}
	\label{zhuguan}
    \vspace{-15pt}
\end{figure*}

In subjective evaluation, we employ GPT-5.2~\cite{achiam2023gpt} as an evaluator, who assesses whether each corrected image contains artifacts related to structure, physics, or distortion. The evaluation scale is binary, with 0 indicating no artifacts and 1 indicating the presence of artifacts. 
Additionally, we perform human evaluation by randomly selecting 50 images from the test set and obtaining evaluations from 20 volunteers. Each volunteer follows the same criteria to rates the images. The results show that our method outperforms other models across all three artifact categories, achieving lower average artifact scores while maintaining high visual realism.
In objective evaluation, we use text-to-image evaluation metrics, with the prompt set uniformly as ``A picture taken by a camera.'' This approach allows us to quantitatively measure the image quality, including perceptual similarity to real-world images, based on various objective scoring functions. Our method achieves the highest scores in HPSv3~\cite{ma2025hpsv3}, CLIP-Score~\cite{taited2023clipscore}, and PickScore~\cite{Kirstain2023Pickscore}, surpassing the second-best results by a significant margin.
Fig.~\ref{zhuguan}(a) presents some subjective results of artifact correction. As seen, our method provides detailed and specific anomaly diagnoses and successfully corrects all artifacts. In contrast, even powerful image editing algorithms such as GPT-Image-1.5 are unable to fully correct all anomalies. 

Noted that in Table~\ref{tab:refine}, ``Ours*'' refers to the results obtained by applying a single-step correction, while ``Ours'' denotes the results from applying multi-step correction. 
The results demonstrate that multi-step iterative correction significantly enhances both the quality and stability of artifact correction, highlighting the effectiveness of the VCoT strategy in progressively refining images and achieving superior outcomes.
Fig.~\ref{zhuguan}(b) shows several examples of multi-step artifact correction. As can be seen, our method progressively eliminates all artifacts through iterative refinement while maintaining the main semantic content of the image.

\begin{table}[t]
    \centering
    \renewcommand{\arraystretch}{1.3}
    \caption{Ablation studies on different training strategies. [Det.: Detection; Corr.: Correction]}
    \label{tab:ablation}
    \vspace{-5pt}
    \resizebox{1.0\linewidth}{!}{
        \begin{tabular}{lccccc}
            \toprule
            \multirow{2}{*}{\textbf{Method}}  & \multicolumn{2}{c}{\textbf{Detection}} & \multicolumn{3}{c}{\textbf{Correction}}\\
            \cmidrule(lr){2-3} \cmidrule(lr){4-6} 
             & Acc. & A.P. & HPSv3 $\uparrow$ & CLIP-Score $\uparrow$ & PickScore $\uparrow$ \\
            
            \midrule
            Only Det. & 96.4 & 98.1 & - & - & -  \\
            Only Corr. & - & - & 5.39 & 21.47 & 18.61 \\
            Only Stage 1 & 97.1 & 98.8 & 5.93 & 21.91 & 18.81 \\
            Only Stage 2 & 97.6 & 99.3 & 5.97 & 21.96 & 18.79 \\
            w/o VCoT & 98.6 & 99.7 & 5.99 & 22.02 & 18.82 \\
            \rowcolor{myblue}
            \textbf{Ours} & 98.8 & 99.8 & 6.20 & 22.12 & 18.86  \\
            \bottomrule
        \end{tabular}
    }
    \vspace{-5pt}
\end{table}

\begin{table}[t!]
    \centering
    \renewcommand{\arraystretch}{1.2}
    \caption{Robustness study of AIGI detection on JPEG compression, Gaussian blur and Resize of GenShield.}
    \label{tab:robust}
    \vspace{-5pt}
    \resizebox{1.0\linewidth}{!}{
    \begin{tabular}{lc|cc|cc|c}
    \toprule[1.5pt]
        \multirow{2}{*}{\textbf{Method}} & \multirow{2}{*}{\textbf{Original}} 
        & \multicolumn{2}{c|}{\textbf{JPEG Compression}}  
        & \multicolumn{2}{c|}{\textbf{Gaussian Blur}} & \textbf{Resize} \\
         & & \textbf{QF=75} & \textbf{QF=70} & \bm{$\sigma = 1.0$} & \bm{$\sigma = 2.0$} & \bm{$\times 0.5$} \\
          \midrule
    {CNNSpot} & 72.9  & 63.5  & 62.4  & 64.5  & 61.7  & 59.9 \\
    {NPR} & 84.0 & 52.2  & 51.6  & 56.8  & 53.4  & 74.3 \\
    {UniFD} & 83.6 &  84.7 & 84.0   &     81.0  &   74.9    &  86.3 \\
    {LaRE} & 85.0  &   62.0    & 63.0      &     54.3  &    54.2   & 51.1 \\
    {AntifakePrompt} & 83.9 & 80.1  & 79.7  & 78.2  & 77.6  & 74.5 \\
    {AIDE} & 97.0  & 92.8  & 92.3  & 91.9  & 90.7  & 89.2 \\
    {RINE} & 96.2  & 92.4  & 91.1  & 94.2  & 92.8  & 92.3 \\
    {AIGI-Holmes} & 95.6 & 93.1 & 92.5 & 92.3 & 90.8 & 90.2  \\
    \midrule
    \rowcolor{myblue}
    \textbf{Ours} & \textbf{98.8} & \textbf{98.0} & \textbf{97.8} & \textbf{97.5} & \textbf{96.9} & \textbf{96.1} \\
    \bottomrule[1.5pt]
    \end{tabular}%
    }
    \vspace{-10pt}
\end{table}

\subsection{Ablation Study}

To evaluate the impact of different training strategies on our model’s performance, we conduct ablation studies with several variants, as shown in Table~\ref{tab:ablation}. The first two rows correspond to training with only detection data and only correction data, respectively. The third and fourth rows report the results of directly training with Stage~1 or Stage~2 alone, without the full curriculum progression. The fifth row removes the multi-step VCoT strategy to examine the contribution of iterative self-correction.

Our results demonstrate that optimizing detection or correction tasks independently, or directly training with only Stage~1 or Stage~2, does not achieve the best performance. By combining both detection and correction tasks in Stage~1 and further refining the model in Stage~2, we obtain the strongest overall results. In addition, removing the multi-round VCoT training data leads to a noticeable degradation in correction quality, highlighting the importance of iterative reasoning for artifact removal. This illustrates the effectiveness of our multi-stage joint optimization strategy, which enables better performance across both detection and artifact correction.



\subsection{Robustness Study of AIGI Detection}
Table~\ref{tab:robust} reports the robustness comparison of AIGI detection methods under these common perturbations. While most baseline detectors suffer noticeable accuracy drops as degradations become stronger, our method remains consistently stable across all settings. In particular, under JPEG compression, we achieve 98.0 and 97.8 accuracy at QF=75 and QF=70, respectively, outperforming the best competing method (AIGI-Holmes, 93.1/92.5) by a large margin. Similar robustness is observed under Gaussian blur, where our accuracy remains 97.5 and 96.9 for $\sigma=1.0$ and $\sigma=2.0$, and under resizing ($\times 0.5$) we still achieve 96.1, consistently higher than all baselines. Overall, these results demonstrate that GenShield provides the most reliable degradation-tolerant detection, which we attribute to the unified understanding–generation training that encourages camera-consistent priors and reduces over-reliance on fragile low-level cues.
\vspace{-10pt}

\section{Conclusion}




We present GenShield, the first unified autoregressive forensics framework that jointly performs explainable AIGI detection and  artifact correction in a curriculum visual CoT mannar, revealing a principled synergy between forensic understanding and generative restoration.
To enable rigorous study of this joint task, we construct GenShield-Set, a large-scale, high-quality dataset with precisely aligned artifact–restored image pairs and a unified evaluation pipeline, filling a critical gap left by detection-only benchmarks. Extensive experiments show that GenShield achieves state-of-the-art performance on mainstream AIGI detection benchmarks while outperforming strong open- and closed-source generators in artifact correction quality and generalization to unseen generators. 

By unifying diagnosis and repair within a single reasoning-driven framework, this work advances AI image forensics from passive verification toward active authenticity restoration, offering a new paradigm for trustworthy generative systems and a meaningful step toward unified understanding–generation modeling in the AI safety community.

\section{Impact Statement}

This paper presents research aimed at advancing methods for AI-generated image forensics, particularly through unified modeling of detection and artifact correction. 
While our work may have potential societal implications related to the analysis of synthetic media, we believe these implications are consistent with those commonly encountered in this area of machine learning research and do not warrant separate discussion here.

\bibliography{example_paper}
\bibliographystyle{icml2026}

\newpage
\appendix
\onecolumn

\section{Limitations}

Despite its effectiveness, GenShield has several limitations. 
First, for images with extremely low generation quality—such as those containing large-scale text corruption, severe geometric distortions, or globally incoherent structures—our method may struggle to recover a fully plausible image within a limited number of iterative correction steps. In such cases, the artifact severity exceeds what can be reliably resolved through gradual diagnose--then--correct refinement.
Second, the quality and diversity of the correction supervision are inherently bounded by the image editing model used during dataset construction. Since the corrected targets are generated by a strong but imperfect editing model, residual biases or failure patterns in this model may be reflected in the training data and consequently limit the upper bound of correction performance. We expect this limitation to be alleviated as more capable image editing models become available.




\section{More Experimental Results}
\label{app:More Experimental Results}

\subsection{Detection Performance on AIGCDetectBenchmark}

We further evaluate our method on the AIGCDetectBenchmark~\cite{zhong2023patchcraft}, which covers a wide range of image generators, including GAN-based models (e.g., ProGAN~\cite{karras2017progressive}, StyleGAN~\cite{Karras2019ASG}, BigGAN~\cite{brock2018large}), diffusion-based models (e.g., Glide~\cite{nichol2021glide}, Stable Diffusion~\cite{rombach2021highresolution}), as well as recent large-scale generators such as DALL·E~2~\cite{ramesh2021zero} and VQDM~\cite{gu2022vector}. This benchmark is particularly challenging due to the large diversity of generation mechanisms and the significant domain gap across generators.

As shown in Table~\ref{table:aigcd}, our method consistently achieves strong detection performance across almost all generators and attains the best overall mean accuracy of 94.93\%, outperforming all competing methods. Notably, our approach maintains near-perfect performance on early GAN-based models such as ProGAN and CycleGAN, while also demonstrating superior robustness on modern diffusion models. For instance, on Glide, our method achieves 98.97\% accuracy, significantly surpassing most existing detectors.
Compared with strong baselines such as AIDE~\cite{yan2024sanity}, PatchCraft~\cite{zhong2023patchcraft}, and AIGI-Holmes~\cite{zhou2025aigi}, our method exhibits more balanced and stable performance across generators. While some baselines achieve high accuracy on specific domains (e.g., DIRE-D on ADM-style diffusion models), they often suffer from substantial degradation on others. In contrast, our method avoids such domain-specific bias and delivers consistently high accuracy across both GAN and diffusion based generators, indicating stronger generalization ability.

\begin{table*}[h]
\caption{\textbf{Detection Performance Comparison on the AIGCDetectBenchmark \cite{zhong2023patchcraft} Benchmark.} Accuracy (\%) of different detectors (rows) in detecting real and fake images from different generators (columns).
The best result and the second-best result are marked in \textbf{bold} and \underline{underline}, respectively.}
\begin{center}
\resizebox{1.0\textwidth}{!}
{%
\begin{tabular}{lccccccccccccccccc}
\toprule
Method  & \rot{ProGAN} &\rot{StyleGAN} & \rot{BigGAN} & \rot{CycleGAN} &\rot{StarGAN} &\rot{GauGAN} &\rot{StyleGAN2} &\rot{WFIR} &\rot{ADM} &\rot{Glide} & \rot{{Midjourney}} & \rot{{SD v1.4}} & \rot{{SD v1.5}}& \rot{{VQDM}}& \rot{{Wukong}}& \rot{{DALLE2}} & \rot{\textit{Mean}} \\ \midrule
CNNSpot 
& \textbf{100.00} & 90.17 & 71.17 & 87.62 & 94.60 & 81.42 & 86.91 & 91.65 & 60.39 & 58.07 & 51.39 & 50.57 & 50.53 & 56.46 & 51.03 & 50.45 & 70.78 \\
FreDect 
& 99.36 & 78.02 & 81.97 & 78.77 & 94.62 & 80.57 & 66.19 & 50.75 & 63.42 & 54.13 & 45.87 & 38.79 & 39.21 & 77.80 & 40.30 & 34.70 & 64.03 \\
Fusing 
& \textbf{100.00} & 85.20 & 77.40 & 87.00 & 97.00 & 77.00 & 83.30 & 66.80 & 49.00 & 57.20 & 52.20 & 51.00 & 51.40 & 55.10 & 51.70 & 52.80 & 68.38 \\
LNP 
& 99.67 & 91.75 & 77.75 & 84.10 & 99.92 & 75.39 & 94.64 & 70.85 & 84.73 & 80.52 & 65.55 & 85.55 & 85.67 & 74.46 & 82.06 & 88.75 & 83.84 \\
LGrad 
& 99.83 & 91.08 & 85.62 & 86.94 & 99.27 & 78.46 & 85.32 & 55.70 & 67.15 & 66.11 & 65.35 & 63.02 & 63.67 & 72.99 & 59.55 & 65.45 & 75.34 \\
UnivFD  
& 99.81 & 84.93 & 95.08 & \underline{98.33} & 95.75 & \textbf{99.47} & 74.96 & 86.90 & 66.87 & 62.46 & 56.13 & 63.66 & 63.49 & 85.31 & 70.93 & 50.75 & 78.43 \\
DIRE-G 
& 95.19 & 83.03 & 70.12 & 74.19 & 95.47 & 67.79 & 75.31 & 58.05 & 75.78 & 71.75 & 58.01 & 49.74 & 49.83 & 53.68 & 54.46 & 66.48 & 68.68 \\
DIRE-D 
& 52.75 & 51.31 & 49.70 & 49.58 & 46.72 & 51.23 & 51.72 & 53.30 & \textbf{98.25} & 92.42 & 89.45 & 91.24 & 91.63 & \underline{91.90} & 90.90 & \underline{92.45} & 71.53 \\
PatchCraft 
& \textbf{100.00} & 92.77 & \underline{95.80} & 70.17 & \textbf{99.97} & 71.58 & 89.55 & 85.80 & 82.17 & 83.79 & \underline{90.12} & \textbf{95.38} & \textbf{95.30} & 88.91 & 91.07 & \textbf{96.60} & 89.31 \\
NPR 
& 99.79 & 97.70 & 84.35 & 96.10 & 99.35 & 82.50 & \underline{98.38} & 65.80 & 69.69 & 78.36 & 77.85 & 78.63 & 78.89 & 78.13 & 76.11 & 64.90 & 82.91 \\
AIDE 
& \underline{99.99} & \textbf{99.64} & 83.95 & \textbf{98.48} & 99.91 & 73.25 & 98.00 & \underline{94.20} & 93.43 & \underline{95.09} & 77.20 & \underline{93.00} & \underline{92.85} & \textbf{95.16} & \textbf{93.55} & \textbf{96.60} & 92.77 \\
AIGI-Holmes 
& \textbf{100.00} & \underline{98.35} & 94.51 & 97.03 & \textbf{100.00} & 95.19 & \textbf{98.88} & \textbf{95.71} & 88.43 & 91.53 & 81.56 & 91.28 & 91.38 & 90.94 & 89.46 & 85.32 & \underline{93.16} \\
\midrule
\rowcolor{myblue}
\textbf{Ours} & \textbf{100.00} & 97.39 & \textbf{96.37} & 97.60 & 99.10 & \underline{98.60} & 97.73 & 94.11 & \underline{94.89} & \textbf{98.97} & \textbf{91.15} & 92.03 & 89.95 & 91.30 & \underline{92.63} & 87.09 & \textbf{94.93} \\
\bottomrule
\end{tabular}
}
\label{table:aigcd}
\end{center}
\end{table*}



\subsection{More Subjective Results}

\begin{figure*}[h]
	\centering
    \includegraphics[width=0.95\linewidth]{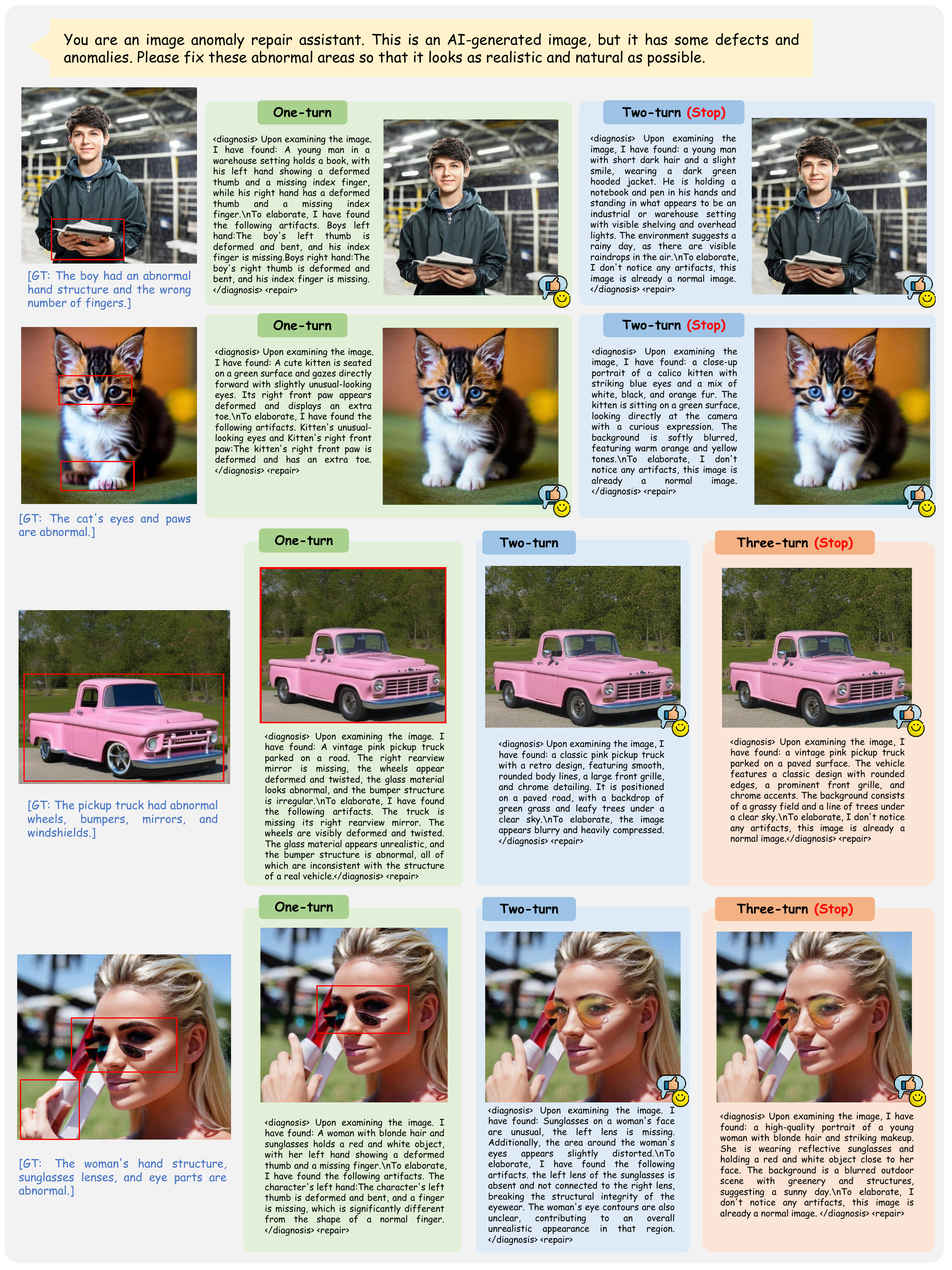}
	\vspace{-5pt}
	\caption{\textbf{Artifact Correction Results of our GenShield.}}
	\label{edit_example1}
    \vspace{-16pt}
\end{figure*}

\begin{figure*}[h]
	\centering
    \includegraphics[width=0.95\linewidth]{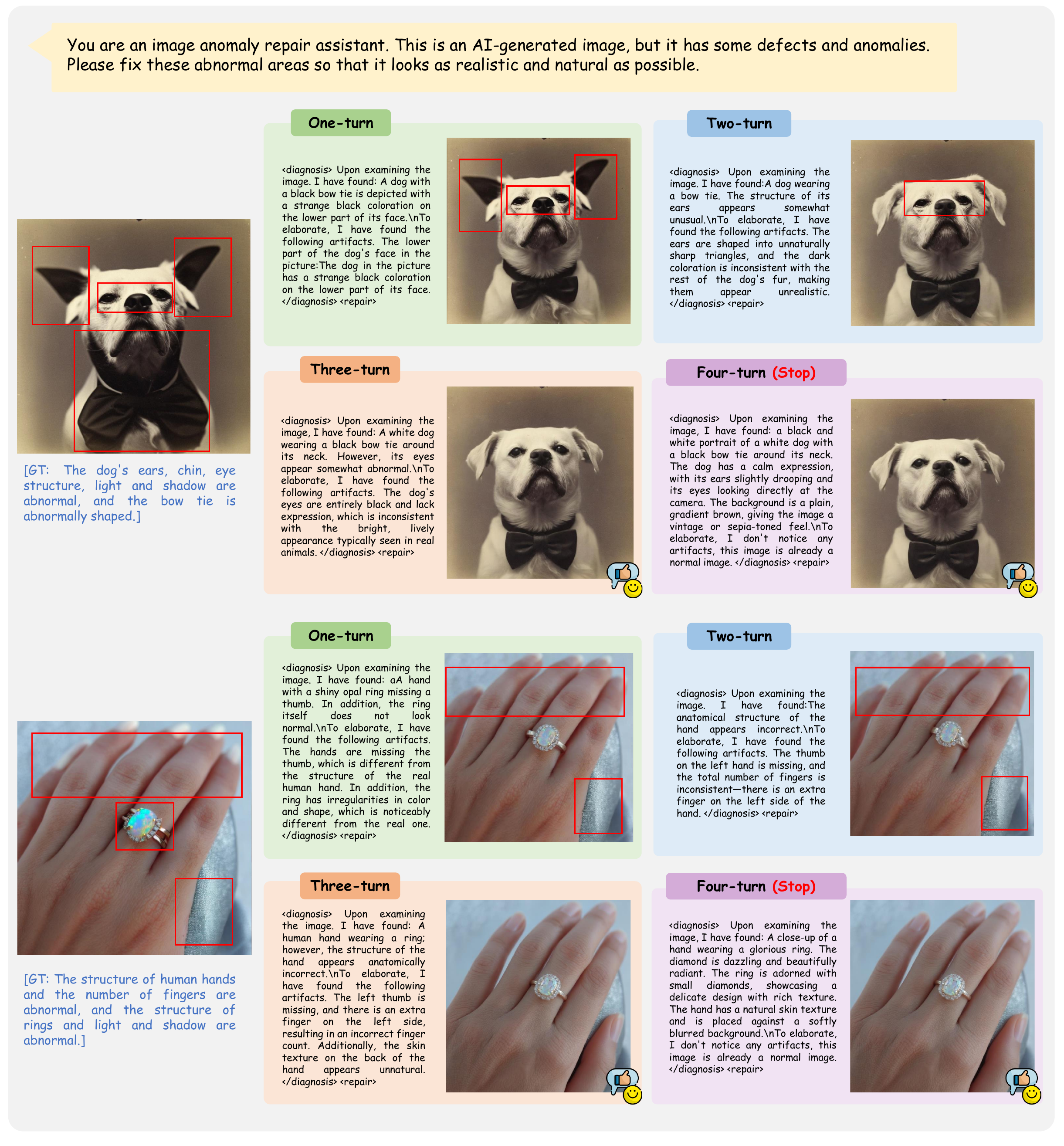}
	\vspace{-5pt}
	\caption{\textbf{More Artifact Correction Results of our GenShield.} }
	\label{edit_example2}
    \vspace{-16pt}
\end{figure*}

\textbf{VCoT Correction Results.}
To better illustrate the effectiveness of our VCoT-based artifact correction, we present representative qualitative examples in Fig.~\ref{edit_example1},~\ref{edit_example2}. As shown, the model performs multi-round, alternating diagnosis and correction: it first describes the suspicious regions in a structured manner and then applies targeted edits to progressively remove artifacts while preserving the main semantics. Importantly, the process is adaptive—when the image is sufficiently corrected, the model outputs a termination-style diagnosis indicating that no obvious artifacts remain, and stops further correction automatically.

\begin{figure*}[h]
	\centering
    \includegraphics[width=0.8\linewidth]{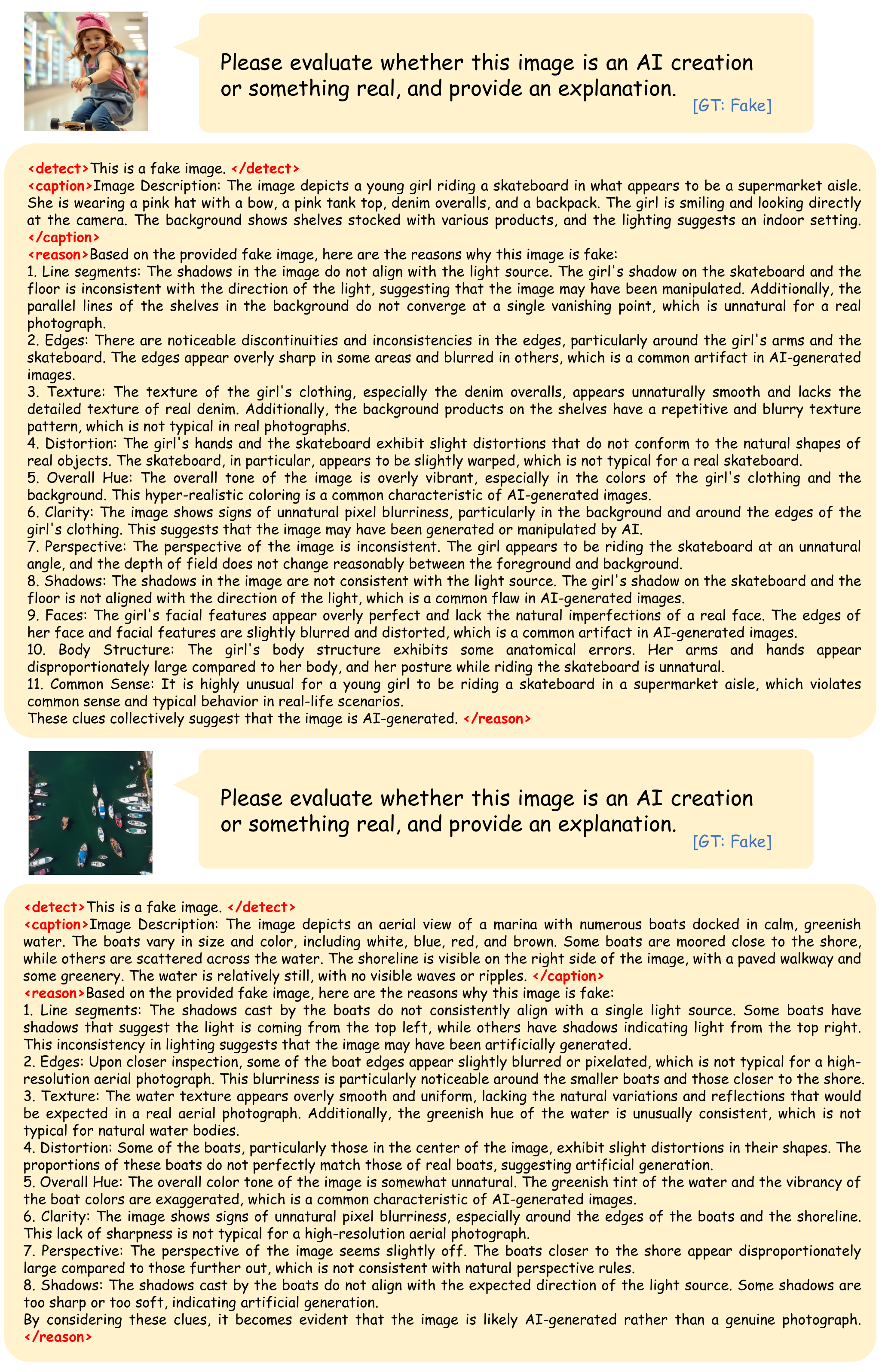}
	\vspace{-5pt}
	\caption{\textbf{AIGI Detection Results of our GenShield.} }
	\label{aigi_fake_example}
    \vspace{-16pt}
\end{figure*}
\begin{figure*}[h]
	\centering
    \includegraphics[width=0.8\linewidth]{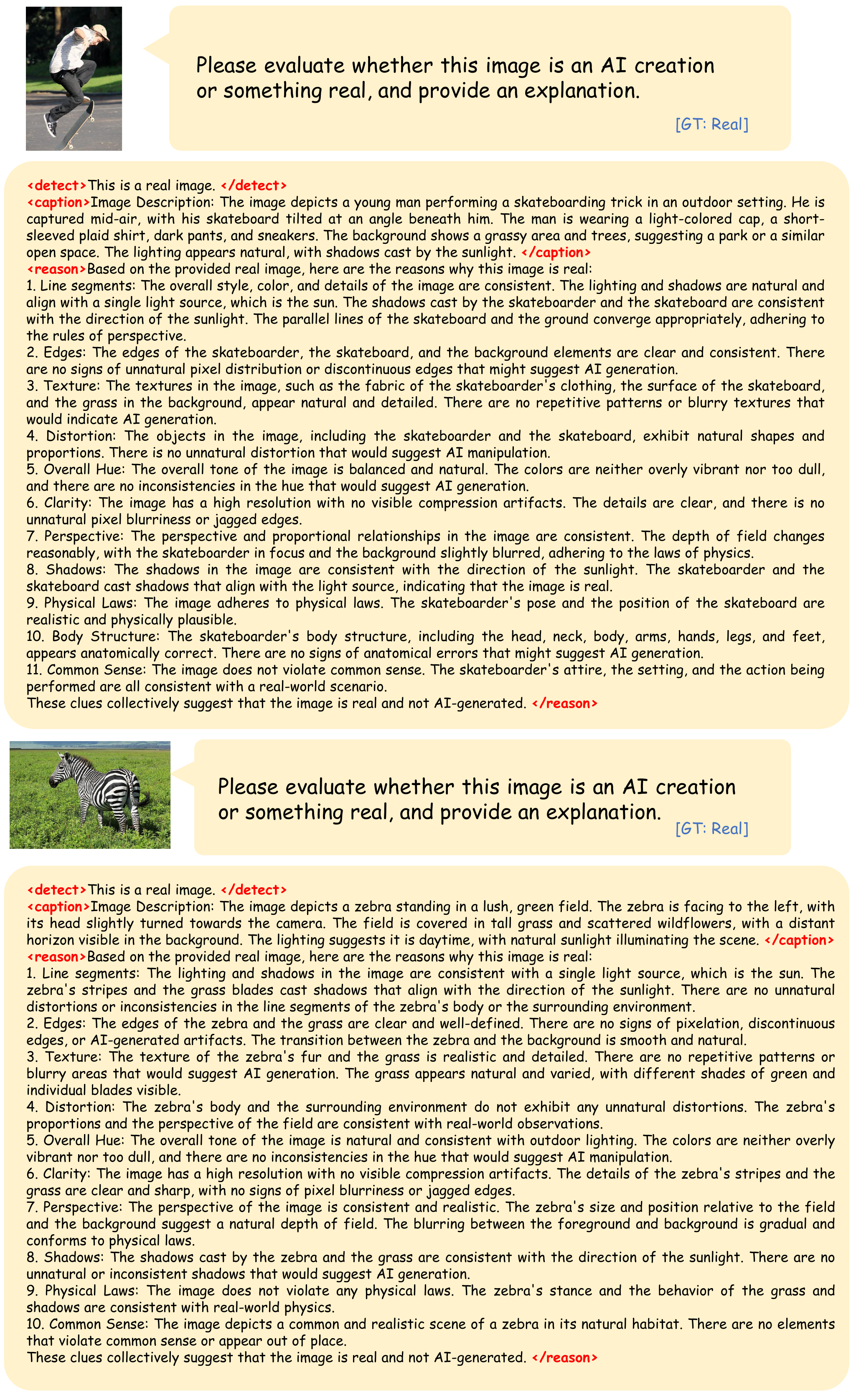}
	\vspace{-5pt}
	\caption{\textbf{More AIGI Detection Results of our GenShield.} }
	\label{aigi_real_example}
    \vspace{-16pt}
\end{figure*}

\textbf{AIGI Detection Results.}
To better demonstrate the effectiveness of our explainable AIGI detection, we present representative qualitative examples in Fig.~\ref{aigi_fake_example},~\ref{aigi_real_example}. As shown, our model outputs a structured prediction together with detailed rationales, analyzing the image from multiple forensic perspectives such as geometry and perspective consistency, lighting and shadow coherence, texture and edge artifacts, and local distortions. These explanations further highlight visually suspicious regions and provide evidence-based cues that support the final real/fake decision.



\subsection{Distribution Analysis of Correction Steps}

\begin{figure}[h]
	\centering
    \includegraphics[width=0.6\linewidth]{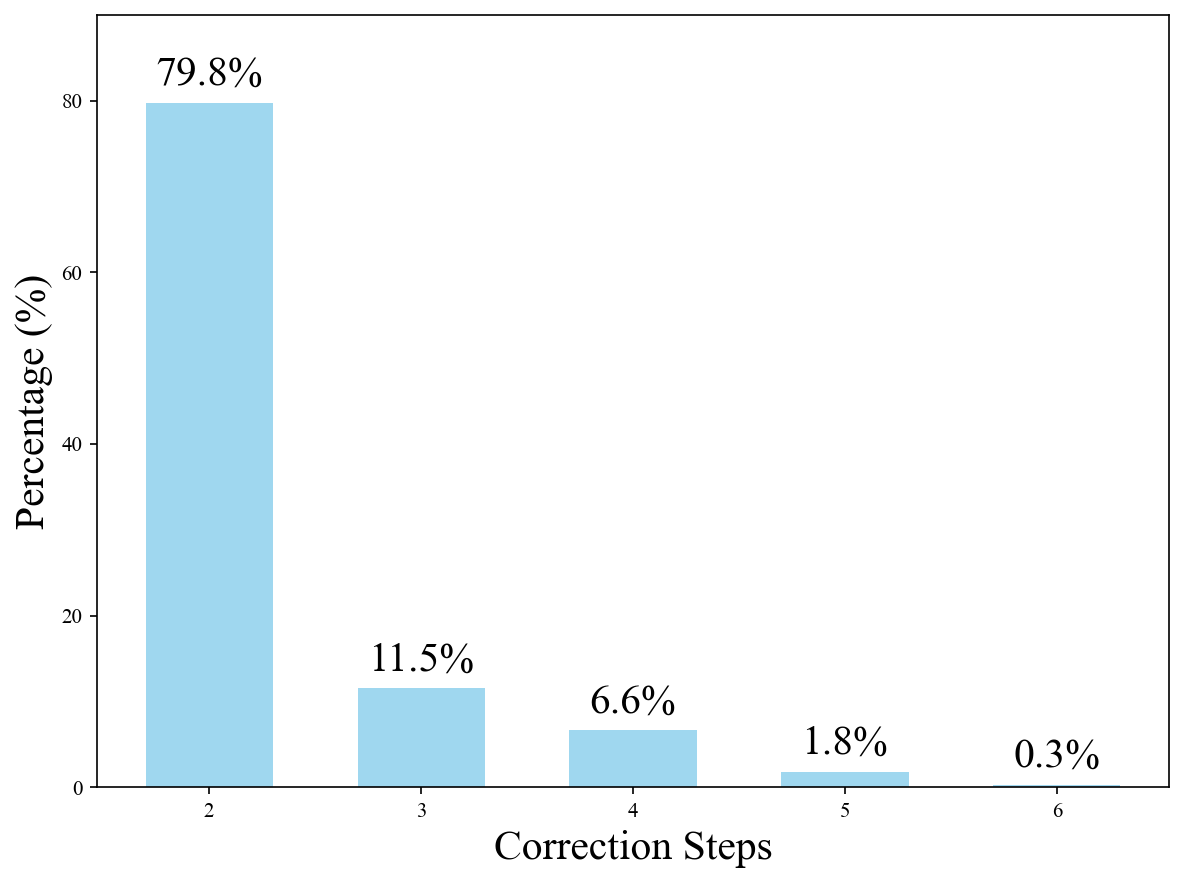}
	\caption{\textbf{Distribution of correction steps on the test set.} Most images are repaired within 2--4 iterative editing rounds, while only a few require more steps.}
    \label{fig:edit_steps}
\end{figure}

Fig.~\ref{fig:edit_steps} shows the distribution of the actual number of correction iterations required in the test set. We observe that the majority of samples can be successfully repaired within only a few editing steps: about 79.8\% of images terminate after two rounds, while 11.5\% require three rounds and 6.6\% require four rounds. Only a very small fraction of difficult cases need more iterations, with 1.8\% finishing in five steps and merely 0.3\% reaching six steps. This distribution demonstrates that our iterative VCoT correction process is both efficient and adaptive, converging quickly for most inputs while still being capable of handling more challenging artifacts through additional refinement.

\section{Details of Artifact Correction Evaluation Pipeline}
\label{app:Details of Artifact Correction Evaluation Pipeline}

Since there is no widely adopted and comprehensive benchmark protocol for evaluating AI image artifact correction, we build a unified evaluation pipeline that measures both artifact removal quality and realistic realism of corrected images. Following the setting in Sec.~\ref{sec:exp_refine}, our pipeline consists of two complementary components: subjective evaluation (artifact-oriented judgments) and objective evaluation (realistic preference and text--image alignment).

\subsection{Subjective Evaluation}
We design a structured, reproducible subjective evaluation protocol to assess whether corrected images still contain common synthetic artifacts. Based on the artifact taxonomy~\cite{kang2025legion} used in our refinement dataset, we categorize artifacts into three groups: Structure (anatomy and geometry), Physics (lighting and interaction), and Distortion (texture and signal). To reduce scoring ambiguity and mitigate evaluator bias introduced by fine-grained rating scales, we adopt a binary scoring rule for each category: 0 indicates no obvious artifacts and 1 indicates artifacts present.

\begin{figure}[t!]
	\centering
    \includegraphics[width=1.0\linewidth]{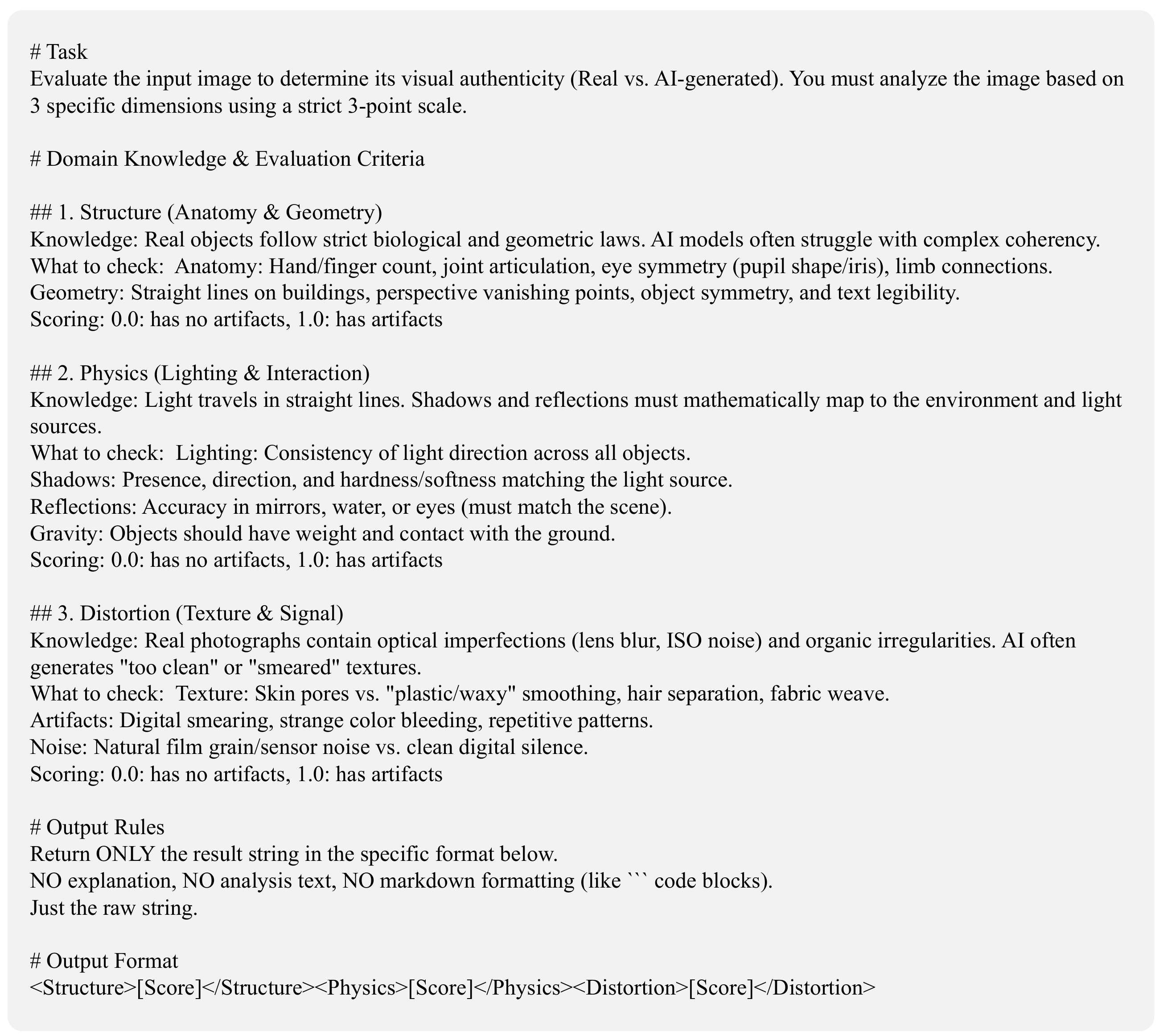}
	\vspace{-5pt}
	\caption{\textbf{GPT-assistant Evaluation Prompt.}}
	\label{gpt_prompt}
    \vspace{-16pt}
\end{figure}

\textbf{GPT-assisted evaluation.}
For scalability and reproducibility, we employ GPT-5.2~\cite{gpt4o} as an automatic evaluator. We prompt the evaluator to judge the authenticity of each corrected image by checking artifacts along the three categories above, and output a structured result with three binary scores. The full prompt template is shown in Fig.~\ref{gpt_prompt}. It provides domain knowledge and explicit criteria for each category (e.g., geometric consistency and text legibility for Structure; shadow/reflection consistency and gravity/contact for Physics; over-smoothed textures, repetitive patterns, and unnatural noise for Distortion), and enforces a strict output format to avoid free-form explanations.

\textbf{Human evaluation.}
To further validate the reliability of GPT-based judgments, we conduct a human study under the same rules. We randomly sample 50 images from the test set and collect ratings for all methods on these samples. We invite 20 volunteers to independently score each corrected result using the same binary criteria for Structure/Physics/Distortion. We report average human scores in Tab.~\ref{tab:refine}, and observe consistent trends with the GPT-assisted evaluation, supporting the robustness of our subjective pipeline.

\subsection{Objective Evaluation}
In addition to subjective judgments, we evaluate corrected images with objective metrics to quantify realistic realism and preference alignment. We follow recent image generation evaluation practice and report HPSv3~\cite{ma2025hpsv3} and PickScore~\cite{Kirstain2023Pickscore}, which are trained on large-scale human preference data, as well as CLIPScore~\cite{taited2023clipscore}, which measures text-image alignment. All three metrics require a text prompt; to avoid method-specific prompt engineering and to ensure a uniform target across all corrected outputs, we fix the prompt to: ``A picture taken by a camera.'' 
This prompt explicitly encodes the desired realistic distribution and provides a consistent reference for comparing different correction methods. Together with the subjective artifact judgments, our pipeline enables a comprehensive and reproducible evaluation of artifact correction quality.

\section{Details of GenShield-Set}
\label{app:Details of GenShield-Set}

As illustrated in Fig.~\ref{dataset}, we build GenShield-Set to support the joint training of explainable detection and synthetic artifact correction. It contains two complementary subsets: \textbf{GenShield-Set-Detect} for detection and rationale generation, and \textbf{GenShield-Set-Correct} for instruction-guided artifact correction and iterative VCoT refinement.

\subsection{GenShield-Set-Correct}

\textbf{Prompt-enhanced correction target generation.}
We construct GenShield-Set-Correct using anomalous images and their anomaly-region textual annotations from SynthScars~\cite{kang2025legion} (Fig.~\ref{dataset}(a)). For each anomalous image, we perform repair prompt enhancement by rewriting the raw region description into a standardized instruction that (i) explicitly specifies the abnormal area and abnormal pattern, and (ii) enforces a realistic, anomaly-free target while preserving identity and global content. 
We then apply Nano Banana Pro~\cite{comanici2025gemini} to generate candidate corrected images.

\textbf{Expert filter for high-quality restorations.}
Automatic correction may produce noisy outputs and failure cases that can bias training. To ensure data quality, we apply an expert-based repaired image filter (Fig.~\ref{dataset}(a)), where 20 trained annotators are instructed to remove candidates with common failure modes, including:
(i) incomplete/failed restoration (artifacts remain visible or correction is insufficient),
(ii) artifact introduction (new unnatural patterns are created),
(iii) content drift (identity/object attributes are changed),
(iv) excessive detail loss (over-smoothing or texture collapse),
(v) structural breakage (geometry/anatomy becomes inconsistent) and
(vi) unnatural seams (local edits mismatch surrounding context).
Each candidate is reviewed by three experts, and we keep it only if at least two experts agree that it is artifact-reduced and semantically consistent with the original. After filtering, we obtain over 10K high-quality corrected images, forming paired supervision tuples for training.

\textbf{Varying repair completeness for multi-step refinement.}
To support multi-step VCoT correction and increase training diversity, we additionally construct varying repair completeness images (Fig.~\ref{dataset}(a)). Specifically, we use our Stage~1 checkpoint to perform instruction-guided correction and retain outputs. These partially corrected images are then treated as the starting point of restoration in Stage~2, enabling the model to learn iterative diagnose--correct behaviors from different correction states rather than only from the original anomalous inputs.

\textbf{Termination-style supervision.}
As shown in Fig.~\ref{dataset}(a), we also create termination-style correction answers for clean or sufficiently corrected images. The model is supervised to explicitly output that no artifacts are observed and the image already appears normal. This supervision is later used to train an explicit stopping behavior in iterative VCoT correction.

\subsection{Correction Data Examples}

\begin{figure*}[h]
	\centering
    \includegraphics[width=0.85\linewidth]{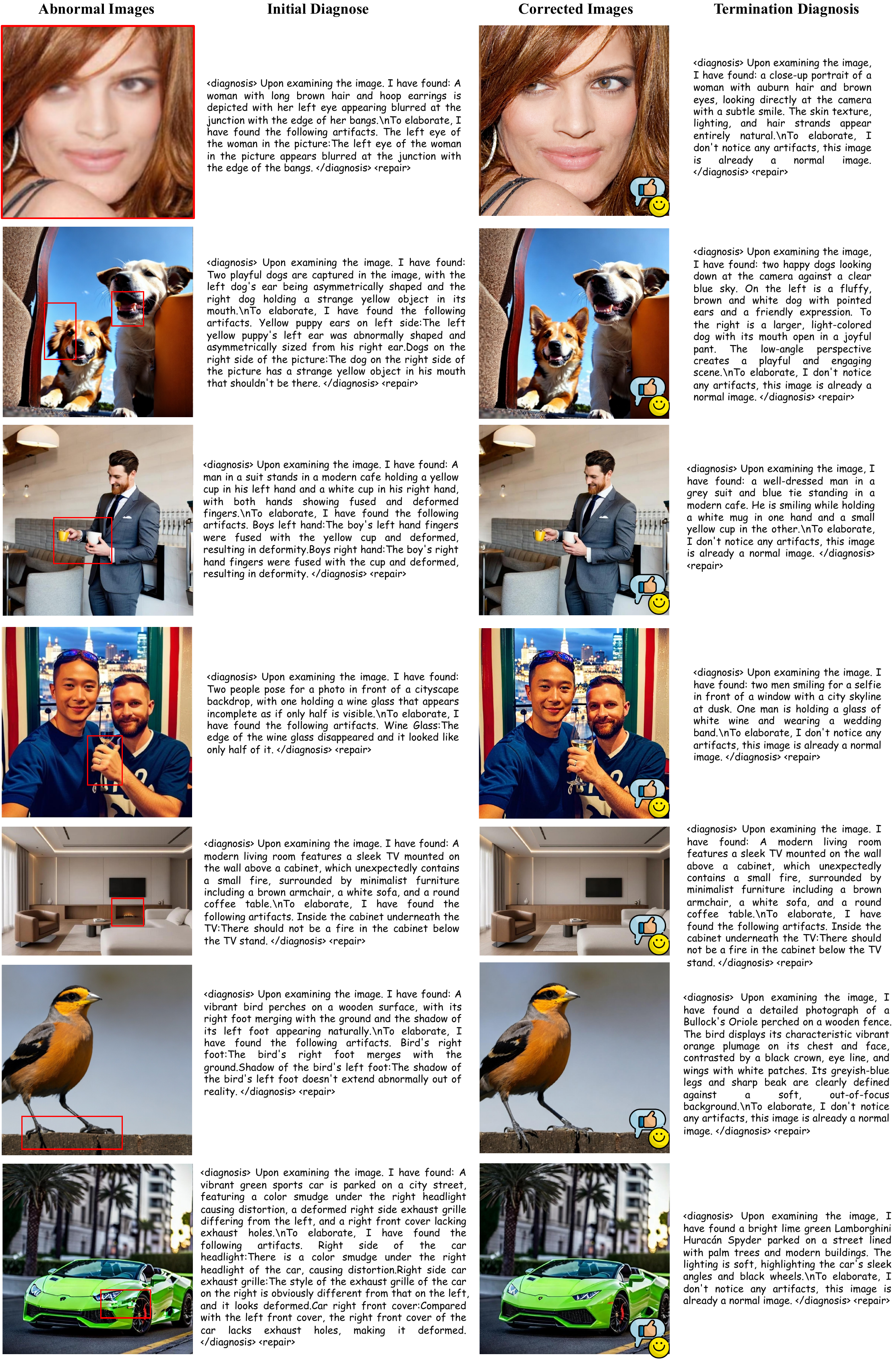}
	\vspace{-5pt}
	\caption{\textbf{Artifact Correction Dataset Samples.}}
	\label{more_dataset}
    \vspace{-16pt}
\end{figure*}

To illustrate the structure of our correction data, we randomly sample several examples from GenShield-Set-Correct, as shown in Fig.~\ref{more_dataset}. Each example contains (i) an anomalous input image $(\textbf{I}_{\text{artifact}}$ with visible artifacts, (ii) an initial diagnostic description $\textbf{T}_{\text{diag}}$ that summarizes the artifact type and indicates the suspicious region, (iii) the corresponding corrected image $\textbf{I}_{\text{correct}}$ after artifact removal, and (iv) a termination diagnosis $\textbf{T}_{\text{stop}}$ generated on the corrected image, where the model explicitly states that no artifacts are observed and the image appears normal. 






\section{More Implementation Details}
\label{app:More Implementation Details}

Table~\ref{tab:training_recipe} summarizes the full training recipe of GenShield, including the two-stage curriculum and the ablation settings (Only Det. and Only Corr.). Across all settings, we adopt a constant learning rate of $2\times10^{-5}$ with zero weight decay, gradient-norm clipping of 1.0, and AdamW optimizer $(\beta_1=0.9,\ \beta_2=0.95,\ \epsilon=1.0\times10^{-15})$. We use 500 warm-up steps for all runs. Stage~1 and Stage~2 are each trained for 5K steps, while the Only Det. and Only Corr. ablations are trained for 10K steps to match the overall update budget. We maintain the same loss weighting between the text cross-entropy loss and the image reconstruction loss with a ratio of $0.25{:}1$ for both Stage~1 and Stage~2. Exponential moving average is applied with EMA ratios of 0.999 for Stage~1 and 0.990 for Stage~2, and 0.990 for both Only Det. and Only Corr. For data preprocessing, correction inputs are randomly resized with the minimum short side and maximum long side in $(512,1024)$, while detection inputs are resized to $(378,980)$. We also apply a diffusion timestep shift of 4.0 for the correction branch. The bottom part of Table~\ref{tab:training_recipe} lists the data sampling ratios used in each setting: Stage~1 jointly samples AIGI detection and instruction-guided artifact correction with a $1.0{:}5.0$ ratio; Stage~2 increases the detection sampling to 2.0 and mixes multiple VCoT correction trajectories with ratios 1.0 (Initial state), 1.0 (Intermediate state), and 0.1 (Terminate state); Only Det. trains solely on detection data; and Only Corr. trains solely on VCoT correction data (Initial state).

\begin{table*}[h!]
\centering

\caption{\textbf{Training recipe of GenShield.}}

\setlength{\tabcolsep}{6pt}
\resizebox{0.9\linewidth}{!}{
\begin{tabular}{l|cccc}
\toprule
 & \textbf{Stage 1} & \textbf{Stage 2} & \textbf{Only Det.} & \textbf{Only Corr.} \\
\hline
\textbf{Hyperparameters} \\
Learning rate  & $2\times10^{-5}$ & $2\times10^{-5}$ & $2\times10^{-5}$ & $2\times10^{-5}$ \\
LR scheduler   &  Constant & Constant & Constant & Constant \\
Weight decay   & 0.0 & 0.0 & 0.0 & 0.0 \\
Gradient norm clip & 1.0 & 1.0 & 1.0 & 1.0  \\
Optimizer       & \multicolumn{4}{c}{AdamW ($\beta_1=0.9$, $\beta_2=0.95$, $\epsilon=1.0 \times 10^{-15}$)} \\
Loss weight (CE : MSE)  & 0.25 : 1 & 0.25 : 1 & - & - \\
Warm-up steps   & 500 & 500 & 500 & 500 \\
Training steps  & 5K & 5K & 10k & 10K \\
EMA ratio       & 0.999 & 0.990 & 0.990 & 0.990 \\
Corr. resolution (min short side, max long side)      & \multicolumn{4}{c}{(512, 1024)} \\
Det. resolution (min short side, max long side)      & \multicolumn{4}{c}{(378, 980)} \\
Diffusion timestep shift & \multicolumn{4}{c}{4.0} \\
\midrule
\textbf{Data sampling ratio} \\
AIGI Detection                              & 1.0 & 2.0 & 1.0 & 0.0 \\
Instruction Guided Artifact Correction      & 5.0 & 0.0 & 0.0 & 0.0 \\
Artifact Correction with VCoT (Initial state)      & 0.0 & 1.0 & 0.0 & 1.0 \\
Artifact Correction with VCoT (Intermediate state)      & 0.0 & 1.0 & 0.0 & 0.0 \\
Artifact Correction with VCoT (Terminate state)      & 0.0 & 0.1 & 0.0 & 0.0 \\
\bottomrule
\end{tabular}}
\label{tab:training_recipe}
\end{table*}

\section{Comparison Methods}

To thoroughly assess AI-generated image detection, we select a broad range of representative baseline methods, spanning frequency domain analyses, gradient-driven detectors, semantic level approaches, reconstruction-based techniques, and others. In this section, we provide detailed descriptions of these comparative methods.

\subsection{AI Synthetic Image Detection Method}

\noindent \textbf{CNNSpot.}~\cite{wang2019cnnspot} CNNSpot builds a forgery detector based on ResNet-50~\cite{he2016resnet} and finds that data augmentations such as JPEG compression and Gaussian blur can effectively improve the model's generalization to unseen generative architectures and data sources.

\noindent \textbf{AntifakePrompt.}~\cite{chang2023antifakeprompt} AntifakePrompt formulates fake image detection as a visual question answering task, leveraging prompt-tuned vision-language models to distinguish real from generated images with zero-shot capability and minimal additional parameters.

\noindent \textbf{UniFD.}~\cite{ojha2023fakedetect} UniFD trains a detector using the feature space extracted by a large pre-trained vision-language model (CLIP: ViT-L/14). The use of a large pre-trained model results in smoother decision boundaries, which enhances the generalization ability of the detector.

\noindent \textbf{NPR.}~\cite{tan2024rethinking} NPR captures and models local pixel dependencies introduced by upsampling operations in CNN-based generative networks to identify universal forgery traces from various generative models, thereby improving the generalization performance of deepfake detection.

\noindent \textbf{LaRE.}~\cite{luo2024lare} LaRE improves diffusion image detection (e.g., DIRE~\cite{wang2023dire}) by using latent reconstruction error as the core feature and introducing an error-guided refinement module (EGRE) to optimize features across spatial and channel dimensions, boosting performance and achieving 8× speedup.

\noindent \textbf{RINE.}~\cite{koutlis2024leveraging} RINE leverages fine-grained features from intermediate CLIP Transformer blocks, maps them into a forgery-aware space via a lightweight network, and uses a learnable module to weight block importance. It is optimized with cross-entropy and contrastive losses, significantly enhancing synthetic image detection and generalization.

\noindent \textbf{AIDE.}~\cite{yan2024sanity} AIDE concatenates DCT-based local frequency features with CLIP global semantics, leveraging multi-expert extraction to capture both low-level artifacts and high-level cues, achieving effective AI-generated image detection and outperforming existing methods on benchmarks.

\noindent \textbf{FreDect.}~\cite{frank2020fredect} FreDect identifies deepfake images by detecting distinct frequency-domain artifacts caused by GAN-generated images. This method leverages abnormal frequency patterns to effectively recognize fake images.

\noindent \textbf{Fusing.}~\cite{ju2022fusing} Fusing presents a two-branch model that combines global and local features to enhance the generalization of AI-synthesized image detection, using multi-head attention to improve the accuracy across various models and image resolutions.

\noindent \textbf{LNP.}~\cite{liu2022detecting} LNP introduces a method for detecting generated images by focusing solely on real images, overcoming common issues in efficiency and generalization, and showing robustness against post-processing with 99.9\% less training data.

\noindent \textbf{LGrad.}~\cite{tan2023learning} The paper introduces a new detection framework called LGrad, which uses gradient learning to detect GAN-generated images by training a transformation model to convert images into a specific gradient space, effectively identifying fake images with reduced training data.

\noindent \textbf{DIRE.}~\cite{wang2023dire} DIRE distinguishes real images from diffusion-generated ones by measuring the error between the input image and its reconstruction from a pre-trained diffusion model, based on the observation that diffusion-generated images can be approximately reconstructed by the diffusion model, whereas real images cannot.

\noindent \textbf{PatchCraft}~\cite{zhong2023patchcraft} PatchCraft introduces an AI-generated image detection method that focuses on texture patches, enhancing the detection of fake images by leveraging the contrast between rich and poor texture regions within an image.

\noindent \textbf{AIGI-Holmes.}~\cite{zhou2025aigi} AIGI-Holmes builds on the Holmes-Set with instruction tuning and human preferences, using a three-stage pipeline that feeds CLIP and NPR features into LLaVa with LoRA fine-tuning. A collaborative decoding strategy fuses vision and semantics, enabling interpretable and generalizable AI-generated image detection.

\noindent \textbf{FakeVLM.}~\cite{wen2025spot} FakeVLM is a multimodal model for synthetic image detection, trained on the 100K-image FakeClue dataset with fine-grained textual annotations. It detects forgeries and generates natural language explanations, achieving high performance and interpretability without extra classifiers.


\subsection{AI Synthetic Image Artifact Correction Method}

\noindent \textbf{GPT-Image-1.5.}~\cite{gpt4o} GPT-Image-1.5 is a flagship image generation model that improves on its predecessor with better prompt understanding, more precise edits, enhanced text rendering, and significantly faster image generation—delivering high-quality, photorealistic outputs in up to ~4× less time.

\noindent \textbf{Nano-Banana.}~\cite{comanici2025gemini} Nano-Banana is a lightweight AI image generation and editing model designed for fast, efficient creative prototyping, supporting multi-round editing consistency and multi-image fusion. With low latency and cost, it is ideal for drafts, social media content, and rapid ideation.

\noindent \textbf{Nano-Banana-Pro.}~\cite{comanici2025gemini} Nano-Banana-Pro is the advanced version built on the Gemini 3 Pro architecture, supporting 4K resolution, accurate multilingual text rendering, and fusion of up to 14 reference images with consistent portrayal of up to 5 individuals. It offers professional-grade controls (e.g., lighting, camera angles), making it suitable for commercial content production and branded assets.

\noindent \textbf{Seedream 4.5.}~\cite{seedream2025seedream} Seedream is an efficient next-generation multimodal image generation system that unifies text-to-image generation, image editing, and multi-image composition through a high-performance Diffusion Transformer + VAE architecture. It can rapidly produce 1K–4K high-resolution images (with 2K image inference in just 1.8 seconds), offering strong generalization and precise multimodal reasoning for complex tasks—making it a powerful interactive tool for both creative and professional applications.

\noindent \textbf{FLUX-Pro.}~\cite{flux2024} FLUX‑Pro is a flagship text-to-image generation model, offering state-of-the-art inference speed, strong prompt adherence, and high visual quality, making it ideal for efficient, scalable AI-based image creation via API.

\noindent \textbf{BAGEL.}~\cite{deng2025emerging} BAGEL is an open-source unified multimodal foundation model that adopts a decoder-only architecture. It is pre-trained on trillions of tokens of interleaved multimodal data, including text, images, and videos, and significantly outperforms other open-source models of the same kind in multimodal understanding/generation and complex reasoning tasks (such as free-form image manipulation, 3D manipulation, etc.).

\noindent \textbf{Qwen-Image-Edit-2511.}~\cite{wu2025qwenimagetechnicalreport} Qwen‑Image‑Edit‑2511 is an enhanced AI image editing model that greatly improves character and multi‑subject consistency over its predecessor, enabling stable natural‑language‑driven edits, identity preservation, and multi‑image fusion. It also integrates built‑in community LoRA capabilities and strengthened geometric reasoning for high‑fidelity, production‑ready editing.

\noindent \textbf{Step1X-Edit-v1p2.}~\cite{liu2025step1x-edit} Step1X-Edit-v1p2 is an advanced open-source model for general image editing, combining a multimodal LLM with a diffusion decoder to generate target images based on reference images and natural language instructions. It achieves state-of-the-art performance on the newly introduced GEdit-Bench, significantly surpassing existing open-source methods and approaching the performance of closed models like GPT-4o and Gemini Flash.

\end{document}